%% file: main.tex
\documentclass[]{worldbench}
\input{preamble}

\title{LiDARCrafter: Dynamic 4D World Modeling \\from LiDAR Sequences}

\author[]{Ao Liang}
\author[]{Youquan Liu}
\author[]{Yu Yang}
\author[]{Dongyue Lu}
\author[]{Linfeng Li}
\author[]{Lingdong Kong~\raisebox{0.15em}{\includegraphics[width=0.017\linewidth]{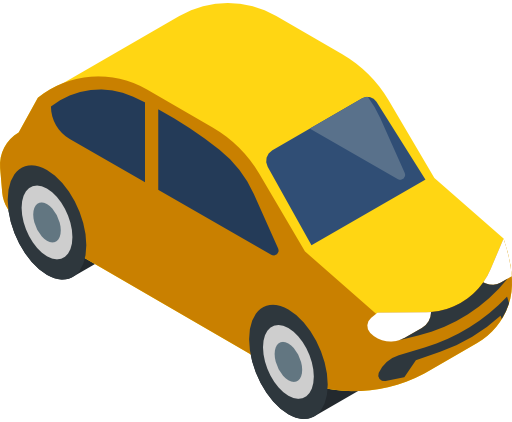}}}
\author[]{Huaici Zhao~\raisebox{0.15em}{\includegraphics[width=0.017\linewidth]{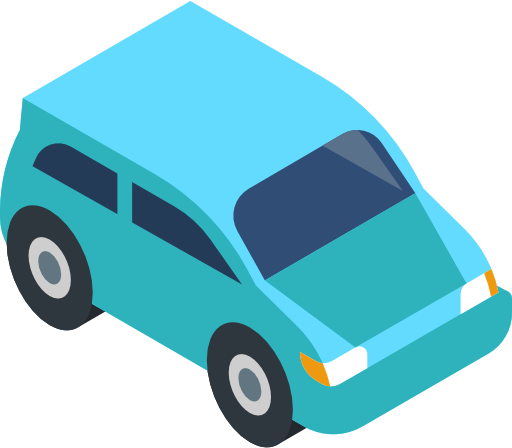}}}
\author[]{\quad\quad\quad\quad\quad Wei Tsang Ooi~\raisebox{0.15em}{\includegraphics[width=0.017\linewidth]{figures/icons/car2.png}}}

\affiliation[]{
\raisebox{-0.1em}{\includegraphics[width=0.029\linewidth]{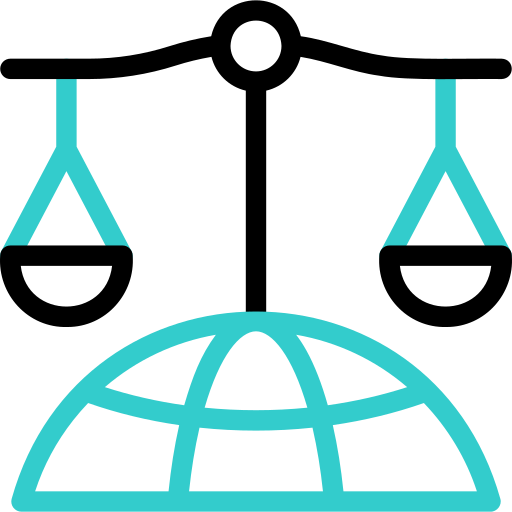}}~WorldBench Team
\\[1.2ex]
~\raisebox{-0.2em}{\includegraphics[width=0.032\linewidth]{figures/icons/car1.png}}~{\small \textbf{Project Lead}}
\quad
\raisebox{-0.2em}{\includegraphics[width=0.031\linewidth]{figures/icons/car2.png}}~{\small \textbf{Corresponding Authors}}
}

\abstract{
Generative world models have become essential data engines for autonomous driving, yet most existing efforts focus on videos or occupancy grids, overlooking the unique LiDAR properties. Extending LiDAR generation to dynamic 4D world modeling presents challenges in controllability, temporal coherence, and evaluation standardization. To this end, we present \textbf{LiDARCrafter}, a unified framework for 4D LiDAR generation and editing. Given free-form natural language inputs, we parse instructions into ego-centric scene graphs, which condition a tri-branch diffusion network to generate object structures, motion trajectories, and geometry. These structured conditions enable diverse and fine-grained scene editing. Additionally, an autoregressive module generates temporally coherent 4D LiDAR sequences with smooth transitions. To support standardized evaluation, we establish a comprehensive benchmark with diverse metrics spanning scene-, object-, and sequence-level aspects. Experiments on the nuScenes dataset using this benchmark demonstrate that LiDARCrafter achieves state-of-the-art performance in fidelity, controllability, and temporal consistency across all levels, paving the way for data augmentation and simulation. The code and benchmark are released to the community.
}

\metadata[
\raisebox{-0.2em}{\includegraphics[width=0.025\linewidth]{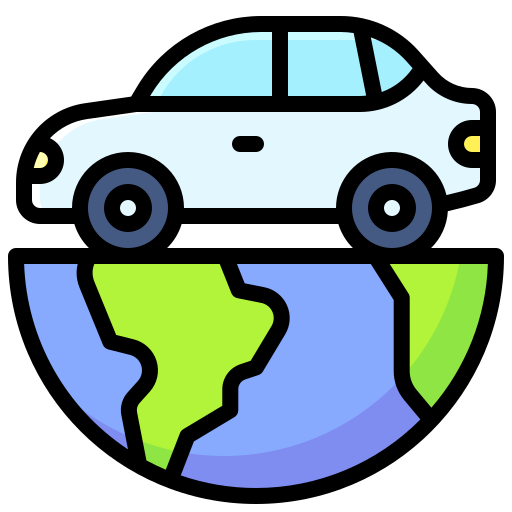}}~~Project Page]{\href{https://lidarcrafter.github.io/}{\texttt{https://lidarcrafter.github.io}}
\\[-1.5ex]}

\metadata[
\raisebox{-0.2em}{\includegraphics[width=0.025\linewidth]{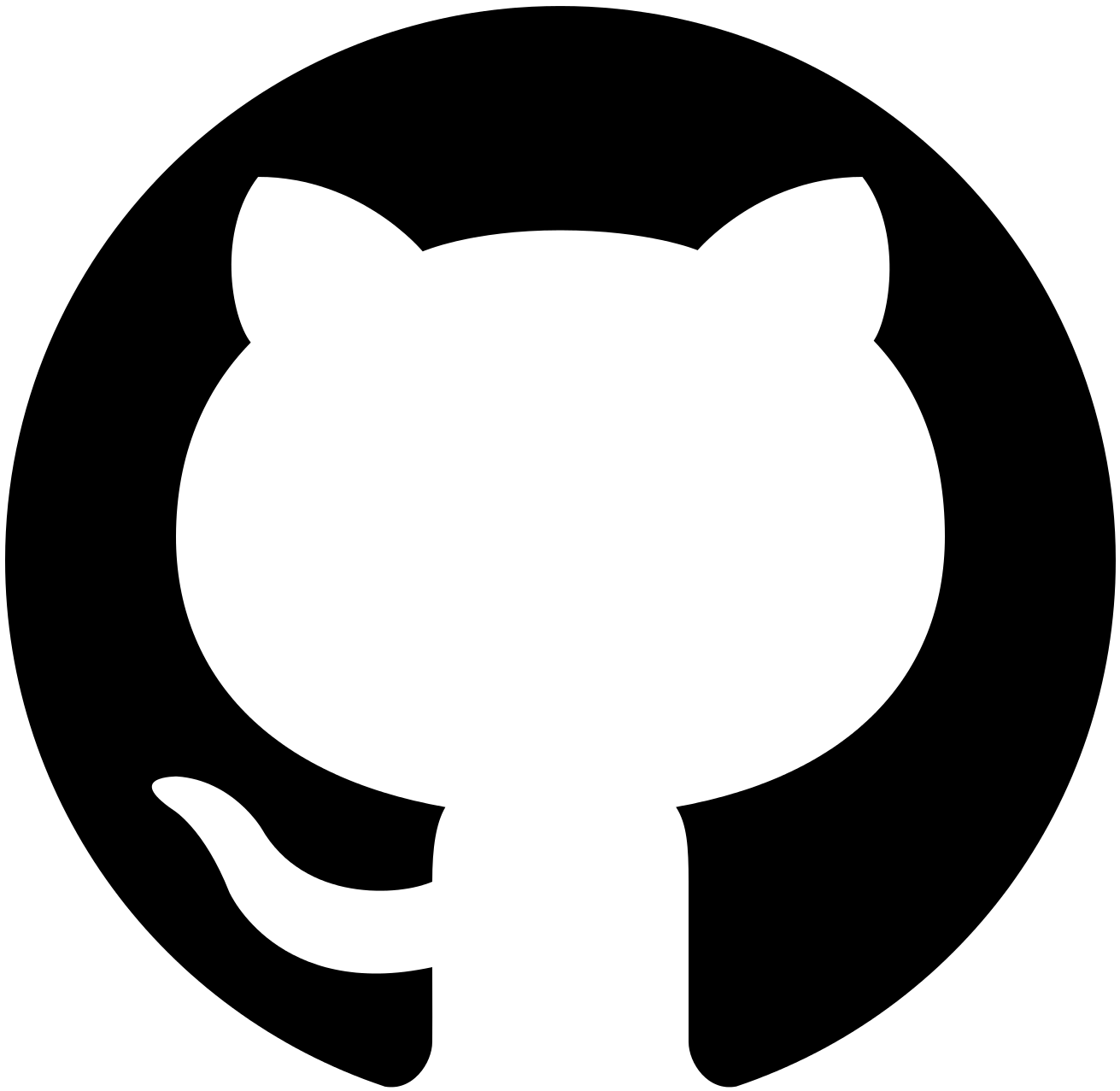}}~~GitHub Repo]{\href{https://github.com/worldbench/lidarcrafter}{\texttt{https://github.com/worldbench/lidarcrafter}}}

\begin{document}

\maketitle

\begin{figure}[h]
    \centering
    \vspace{0.2cm}
    \includegraphics[width=\textwidth]{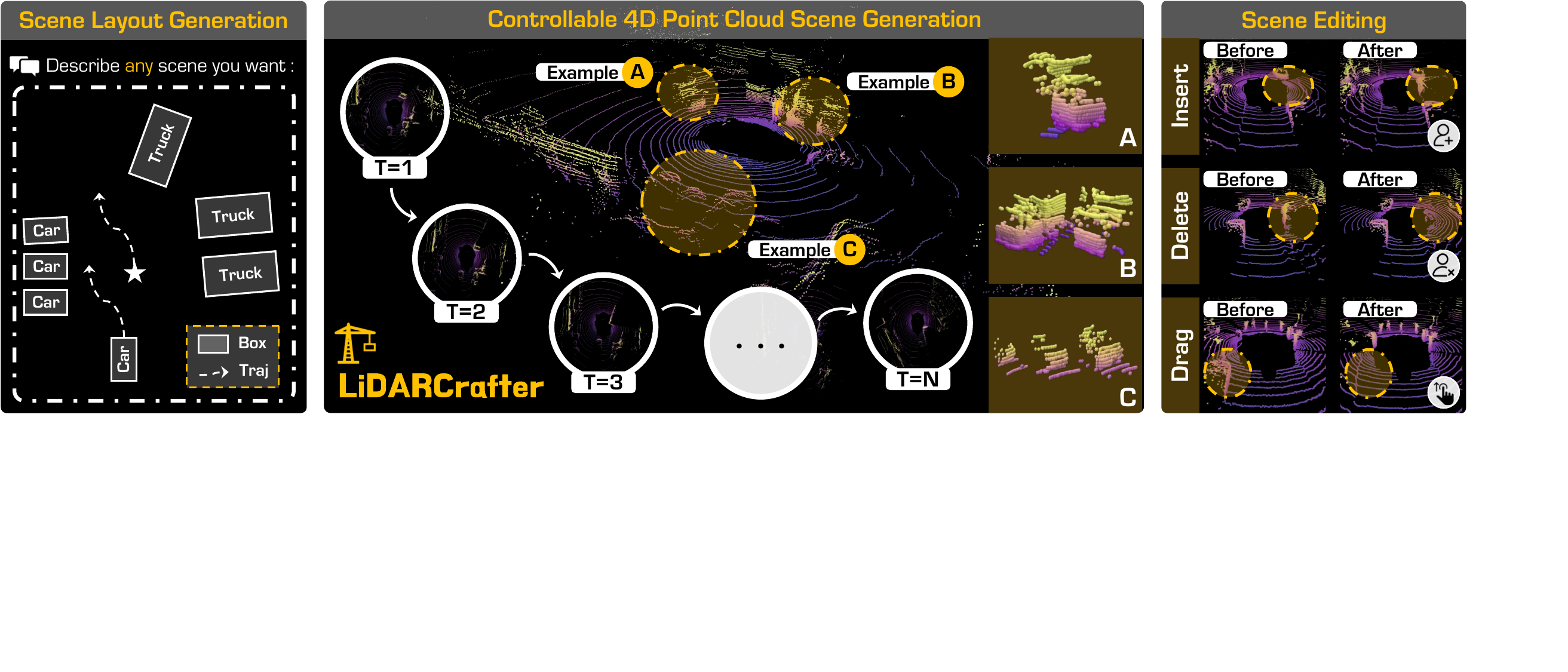}
    \vspace{-0.6cm}
    \caption{We propose \textbf{LiDARCrafter}, a 4D LiDAR-based generative world model that supports controllable point cloud layout generation (\textbf{left}), dynamic sequential scene generation (\textbf{center}), and rich scene editing applications (\textbf{right}). Our framework enables intuitive \emph{``what you describe is what you get''} LiDAR-based 4D world modeling.}
    \label{fig:teaser}
    \vspace{-0.2cm}
\end{figure}

\input{sections/1_intro}
\input{sections/2_related_work}
\input{sections/3_method}
\input{sections/4_experiments}
\input{sections/5_conclusion}

\vspace{0.2cm}
\beginappendix

\startcontents[appendices]
\printcontents[appendices]{l}{1}{\setcounter{tocdepth}{2}}
\vspace{0.2cm}

\input{supp_sections/1_additional_details}
\input{supp_sections/2_addtional_exp}
\input{supp_sections/3_impact}
\clearpage\clearpage
\input{supp_sections/4_ack}

\clearpage\clearpage
\bibliographystyle{plainnat}
\bibliography{main}

\end{document}

%% file: preamble.tex
\definecolor{w_blue}{RGB}{52,204,204}
\definecolor{w_yellow}{RGB}{255,192,0}

\usepackage{amsfonts}
\usepackage{amsmath}
\usepackage{amssymb}
\usepackage{algorithm}
\usepackage{algorithmic}

\usepackage{colortbl}

\usepackage{makecell}
\usepackage{multicol}
\usepackage{multirow}

\usepackage{pifont}
\newcommand{\cmark}{\ding{51}}
\newcommand{\xmark}{\ding{55}}

\usepackage{wrapfig}

\definecolor{crafter}{RGB}{208,159,18}
\newcommand{\ours}{\textbf{LiDARCrafter}}

\makeatletter
\def\blfootnote{\xdef\@thefnmark{}\@footnotetext}
\DeclareRobustCommand\onedot{\futurelet\@let@token\@onedot}
\def\@onedot{\ifx\@let@token.\else.\null\fi\xspace}

\makeatother

\def\eqref#1{Equation~\ref{#1}}

\usepackage{soul}

\usepackage{titletoc}

%% file: sections/1_intro.tex
\section{Introduction}
\label{sec:intro}

Generative world models are rapidly advancing the synthesis and interpretation of large-scale sensor data for autonomous driving~\cite{survey_3d_4d_world_models,hu2023gaia,mei2024dreamforge,yan2025adr1}. Recent work has concentrated on structured modalities such as video and occupancy grids, whose dense, regular structure fits image and voxel pipelines~\cite{gao2023magicdrive,wang2024occsora}. LiDAR, though indispensable for metric 3D geometry and all-weather perception, remains underexplored. The point clouds acquired by LiDAR sensors are often sparse, unordered, and irregular \cite{kong2023robo3d,liang2025pi3det,liu2023seal,xu2024superflow,xie2025drivebench,liu2025lalalidar,li2025_3eed,zhu2025spiral}. Therefore, techniques designed for images or grids do not transfer cleanly.

Early efforts, such as LiDARGen, project full $360^{\circ}$ scans to range images and borrow pixel-based methods \cite{zyrianov2022learning}. Later approaches improve single-frame fidelity but stop short of dynamic sequences \cite{nakashima2024lidar,ran2024towards}. Multimodal systems like UniScene and GENESIS rely on occupancy or video as intermediaries, limiting LiDAR independence and increasing computation~\cite{li2025uniscene, guo2025genesis}. A dedicated 4D LiDAR world model is therefore still absent.

Bridging this gap requires progress on three fronts. First, \emph{user controllability}. Text prompts are the most accessible interface, yet they lack the spatial detail to specify LiDAR scenes, whereas precise inputs such as 3D boxes, HD maps, or trajectories demand costly annotation \cite{hu2023gaia,bian2024dynamiccity}. Second, \emph{temporal consistency}. A single frame cannot reveal occlusion patterns or object kinematics for downstream utilization, necessitating a full 4D generation model. Conventional cross-frame attention strategies neglect the geometric continuity inherent in point clouds. Third, \emph{standardized evaluation}. Existing video world models benefit from established benchmarks that score visual quality, perception accuracy, and closed-loop applicability \cite{huang2024vbench}. LiDAR still lacks standard protocols for evaluating fidelity and consistency across perspectives.

To close these gaps, we introduce \ours{}, a unified framework for controllable 4D LiDAR sequence generation. The key is an explicit, object-centric 4D layout that captures geometry and motion while achieving \emph{accessible yet precise control} (see Figure~\ref{fig:teaser}). In \textbf{Text2Layout}, a large language model (LLM) lifts the user prompt to an ego-centric scene graph, and a tri-branch diffusion network predicts object boxes, trajectories, and shape priors, yielding the full 4D layout. \textbf{Layout2Scene} feeds the layout to a range-image diffusion network and produces a high-fidelity first scan. The explicit layout modeling also enables fine-grained control like insertion, deletion, and dragging. \textbf{Scene2Seq} then generates the remaining frames \emph{autoregressively}, warping past points with motion priors to keep temporal coherence. To standardize 4D LiDAR world models assessment, we release an \emph{evaluation suite} that scores scene, object, and sequence quality.

Experiments on the nuScenes dataset~\cite{caesar2020nuscenes} demonstrate that our method achieves top performance in both single-frame fidelity and sequence coherence, while offering intuitive control and editing capabilities. This work establishes a new benchmark for LiDAR-based dynamic world modeling in autonomous driving.  

In summary, the core contributions of this work are: 
\begin{itemize}
    \item We present \ours{}, the \textbf{first} 4D generative world model dedicated to LiDAR data, with superior controllability, spatiotemporal consistency, and autoregressive generation capability.
    \item We introduce a tri-branch 4D layout conditioned pipeline that turns language into an editable 4D layout and uses it to guide temporally stable LiDAR sequence synthesis.
    \item We propose a comprehensive \textbf{Evaluation Suite} for LiDAR sequence generation, encompassing scene-level, object-level, and sequence-level metrics.
    \item We demonstrate \textbf{best} single-frame and sequence-level LiDAR point cloud generation performance on nuScenes, with improved foreground quality over existing methods.
\end{itemize}

\begin{figure}[t]
    \centering
    \includegraphics[width=\textwidth]{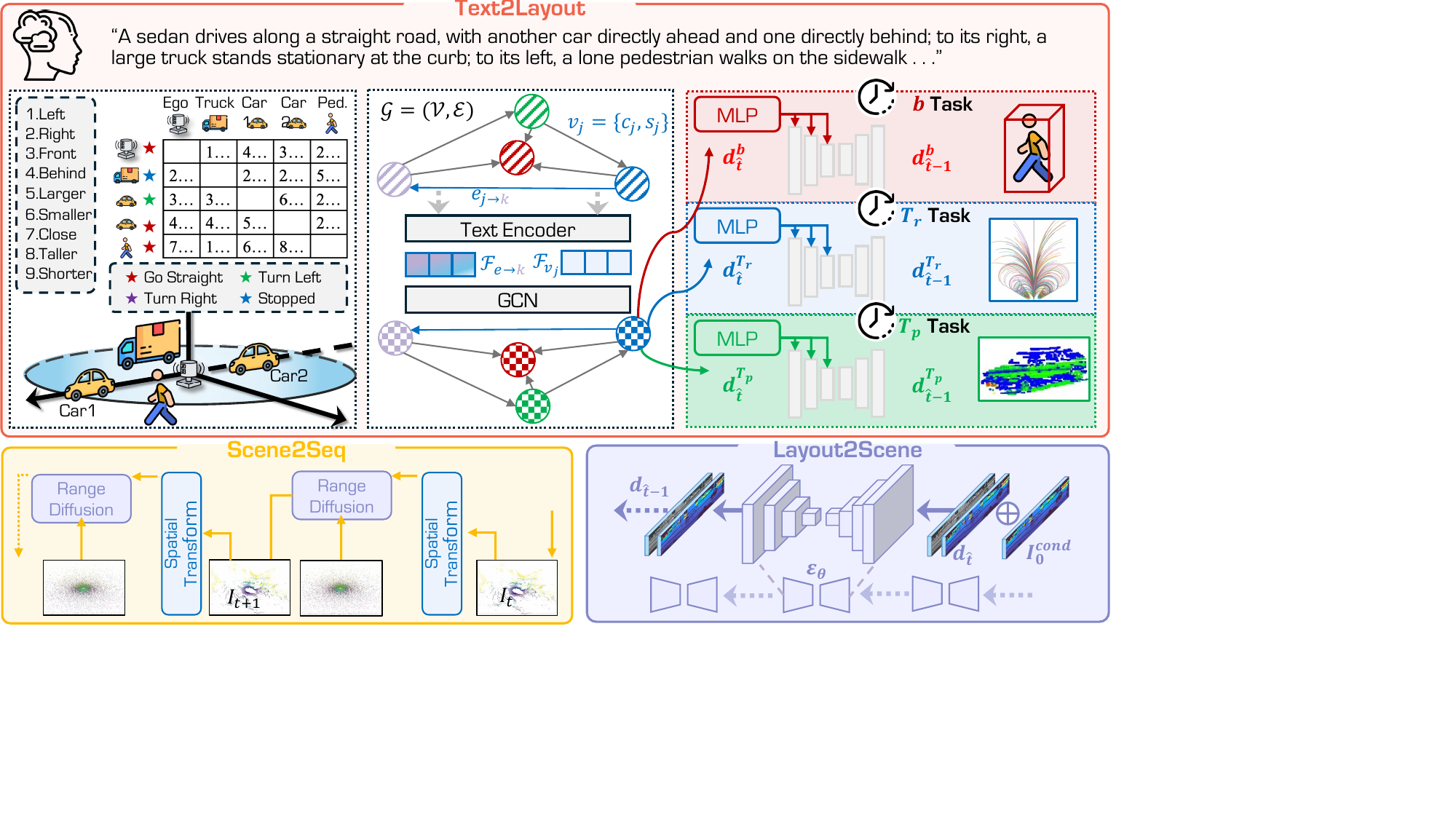}
    \vspace{-0.5cm}
    \caption{
    Framework of \ours{}. In the \textbf{Text2Layout} stage (\emph{cf.}~Section~\ref{sec:layout_generation}), the natural-language instruction is parsed into an ego-centric scene graph, and a tri-branch diffusion network generates 4D conditions for bounding boxes, future trajectories, and object point clouds. In the \textbf{Layout2Scene} stage (\emph{cf.}~Section~\ref{sec:static_pointcloud_generation}), a range-image diffusion model uses these conditions to generate a static LiDAR frame. In the \textbf{Scene2Seq} stage (\emph{cf.}~Section~\ref{sec:4D_pointcloud_generation}), an autoregressive module warps historical points with ego and object motion priors to generate each subsequent frame, producing a temporally coherent LiDAR sequence.
    }
    \label{fig:framework}
\end{figure}

%% file: sections/2_related_work.tex
\section{Related Work}
\label{sec:related}

\noindent\textbf{Driving Generative World Models.}  
Generative world models aim to simulate environmental dynamics to aid autonomous driving decisions. Recent works primarily use video and occupancy representations. Video-based methods like GAIA-1~\cite{hu2023gaia}, DreamForge~\cite{mei2024dreamforge}, and Epona~\cite{zhang2025epona} employ autoregressive modeling and richer conditioning for improved synthesis. MagicDrive~\cite{gao2023magicdrive} utilizes a bird's-eye view (BEV) for temporal consistency~\cite{wang2024drivedreamer,zhao2025drivedreamer,ni2025maskgwm}. Occupancy-based methods, such as OccWorld~\cite{zheng2024occworld}, GaussianWorld~\cite{zuo2025gaussianworld}, QuadricFormer~\cite{zuo2025quadricformer}, OccSora~\cite{wang2024occsora}, and DynamicCity~\cite{bian2024dynamiccity}, provide structured spatial representations beneficial for downstream tasks. Multimodal models like UniScene~\cite{li2025uniscene} and GENESIS~\cite{guo2025genesis} align cross-modal features for consistency. Despite its importance, LiDAR remains underexplored, mostly limited to forecasting-oriented methods~\cite{liu2025towards,zhang2023copilot4d}.

\noindent\textbf{LiDAR Point Cloud Generation.}  
Early LiDAR generative methods, including LiDARGen~\cite{zyrianov2022learning}, projected LiDAR point clouds into range images. Subsequent approaches like RangeLDM~\cite{hu2024rangeldm}, Text2LiDAR~\cite{wu2024text2lidar}, and WeatherGen~\cite{wu2025weathergen} introduced latent diffusion models~\cite{rombach2022high} or diverse conditioning signals. R2DM~\cite{nakashima2024lidar} and R2Flow~\cite{nakashima2024fast} adopted single-stage diffusion models~\cite{ho2020denoising} for geometric precision. UltraLiDAR~\cite{xiong2023ultralidar} utilized BEV-based VQ-VAE~\cite{van2017neural} for rich scene editing, further extended by OpenDWM~\cite{opendwm}. X-Drive~\cite{xie2024x} and UniScene~\cite{li2025uniscene} leveraged cross-modal consistency. Yet, dynamic 4D sequence generation and fine-grained controllability remain an unresolved problem.

\noindent\textbf{Controllability in Scene Synthesis.}  
Fine-grained control is crucial in generative modeling. Current methods depend on detailed inputs such as BEV semantic maps~\cite{gao2023magicdrive,mei2024dreamforge}, HD maps~\cite{swerdlow2024street}, and 3D bounding boxes~\cite{zhang2024perldiff,yang2024drivearena}, but preparing these inputs is labor-intensive. Text-based methods~\cite{wu2024text2lidar,hu2023gaia,mei2024dreamforge} offer simpler control but typically lack spatial precision. Two-stage indoor scene synthesis methods~\cite{zhai2024echoscene,yang2025mmgdreamer} use intermediate scene graphs to improve spatial control, recently extended to outdoor static occupancy grids~\cite{liu2025controllable,yang2025x}. Mature methods for controllable generation and editing of dynamic 4D LiDAR scenes, however, are not yet established.

%% file: sections/3_method.tex
\section{LiDARCrafter: 4D LiDAR World Model}
\label{sec:method}

We propose \textbf{LiDARCrafter}, the first LiDAR world model that turns free-form user instructions into temporally coherent 4D point cloud sequences with object-level control.

The cornerstone is an explicit 4D foreground layout that bridges the descriptive power of language and the geometric rigor required by LiDAR. As shown in Figure~\ref{fig:framework}, our framework adopts a three-stage process. In the \textbf{Text2Layout} stage (\emph{cf.}~Section~\ref{sec:layout_generation}), an LLM converts the instruction into an ego-centric scene graph, and a tri-branch diffusion sampler generates \emph{object boxes, trajectories, and shape priors}, which serve as the conditioning layout signal. In the \textbf{Layout2Scene} stage (\emph{cf.}~Section~\ref{sec:static_pointcloud_generation}), a range-image diffusion model turns the layout into a high-fidelity static scan. In the \textbf{Scene2Seq} stage (\emph{cf.}~Section~\ref{sec:static_pointcloud_generation}), the static cloud is autoregressively warped and inpainted to yield drift-free frames. Finally, our \textbf{EvalSuite} (\emph{cf.}~Section~\ref{sec:evaluation}) adds metrics for object semantics, layout soundness, and motion fidelity, giving the first comprehensive benchmark for 4D LiDAR generation.

\subsection{Text2Layout: 4D Layout Generation}
\label{sec:layout_generation}
Natural-language prompts alone lack the spatial precision needed for complex world modeling. We therefore introduce a scene graph as an intermediate, explicit encoding of object geometry and relations. LLMs can parse text into such graphs, a strategy proven effective for scene synthesis \cite{feng2023layoutgpt,yang2025mmgdreamer}. 
\textbf{LiDARCrafter} extends this idea to dynamic outdoor settings. The LLM first builds a 4D scene graph from the prompt. A diffusion decoder then transforms this graph into a detailed layout of object boxes, trajectories, and shape priors that guides the subsequent LiDAR sequence generation process.

\vspace{0.5mm}
\noindent\textbf{Language-Driven Graph Construction.}
Given a textual instruction, we prompt an LLM~\cite{achiam2023gpt} to build an ego-centric scene graph \(\mathcal{G} = (\mathcal{V}, \mathcal{E})\).
A tailored query enumerates all foreground objects, producing the node set
\(\mathcal{V} = \{v_0, \dots, v_M\}\), where \(v_0\) denotes the ego vehicle and the remaining \(M\) nodes represent dynamic objects.
Each node \(v_i\) is annotated with its semantic class \(c_i\) and a motion state phrase \(s_i\) (\emph{e.g.}, ``go straight''). For every ordered pair \((i, j)\) with \(i \neq j\), a directed edge \(e_{i \to j} \in \mathcal{E}\) encodes their spatial relation (\emph{e.g.}, ``in front of'', ``lager than'', details in Figure~\ref{fig:framework}).
Unlike prior work~\cite{liu2025controllable}, including the ego node yields a structurally complete scene graph that fully conditions downstream layout generation.

\vspace{0.5mm}
\noindent\textbf{Scene-Graph Lifting.}
Given a textual scene graph, we aim to infer for each node \(v_i\) a 4D layout tuple $\mathcal{O}_i=\bigl(\mathbf{b}_i,\boldsymbol{\delta}_i,\mathbf{p}_i\bigr),$ where \(\mathbf{b}_i=(x_i,y_i,z_i,w_i,l_i,h_i,\psi_i)\) is the 3D bounding box capturing the 3D center, size, and yaw angle of object \(i\). \(\boldsymbol{\delta}_i=\{(\Delta x_i^{\,t},\Delta y_i^{\,t})\}_{t=1}^{T}\) records planar displacements over \(T\) future frames, and \(\mathbf{p}_i\in\mathbb{R}^{N\times3}\) stores \(N\) canonical foreground points that sketch the shape of object. This tuple captures where, how, and what for every node and serves as the target of our denoiser during the diffusion process.

\vspace{0.5mm}
\noindent\textbf{Graph-Fusion Encoder.}
To obtain context-aware priors for every tuple, following the method in the indoor area~\cite{zhai2024echoscene}, we process the scene graph \(\mathcal{G}=(\mathcal{V},\mathcal{E})\) with an \(L\)-layer TripletGCN~\cite{johnson2018image}. We first embed nodes and edges with a frozen CLIP text encoder~\cite{radford2021learning} to bring richer semantics:
\begin{equation}
\begin{aligned}
&\mathbf{h}_{v_i}^{(0)} = \texttt{concat}(\text{CLIP}(c_i),\,\text{CLIP}(s_i),\,\boldsymbol{\omega}_i), \\
&\mathbf{h}_{e_{i\to j}}^{(0)} = \text{CLIP}(e_{i\to j}),
\end{aligned}
\end{equation}
where \(\boldsymbol{\omega}_i\) is a learnable positional code. At layer \(\ell\), we update triplets with two lightweight MLPs:
\(\Phi_{\text{edge}}\) for edge reasoning and \(\Phi_\text{agg}\) for neighborhood aggregation as follows:
\begin{equation}
\begin{aligned}
&({\tilde{\mathbf{h}}}_{v_{i}}^{(\ell)}, \mathbf{h}_{e_{i\to j}}^{(\ell+1)}, \tilde{\mathbf{h}}_{v_{j}}^{(\ell)}) = \Phi_{\text{edge}}(\mathbf{h}_{v_{i}}^{(\ell)}, \mathbf{h}_{e_{i\to j}}^{(\ell)}, \mathbf{h}_{v_{j}}^{(\ell)}), \\
&\mathbf{h}_{v_{i}}^{(\ell+1)} = \tilde{\mathbf{h}}_{v_{i}}^{(\ell)} + \Phi_{\text{agg}}\Big( \texttt{avg}\big( \tilde{\mathbf{h}}_{{v}_{j}}^{(\ell)}~|~v_j \in N_{\mathcal{G}}(v_{i}) \big) \Big).
\end{aligned}
\label{eq:gcn}
\end{equation}
After \(L\) hops, each node feature \(\mathbf{h}^{(L)}_{v_i}\) encodes both global semantics and local geometry, providing a strong semantic-geometric prior for LiDAR layout generation.

\vspace{0.5mm}
\noindent\textbf{Layout Diffusion Decoder.}
The final node embeddings condition a tri-branch diffusion decoder~\cite{rombach2022high}, one branch per element of \(\mathcal{O}_i\). Let \(\mathbf{d}^o_\tau\) be the noisy sample of modality \(o\in\mathcal{O}_i\) at timestep \(\tau\). Each branch minimizes:
\begin{equation}
\mathcal{L}^{o}=\mathbb{E}_{\tau,\mathbf{d}^o,\varepsilon}
\bigl\lVert\varepsilon-\varepsilon^{o}_{\theta}(\mathbf{d}^o_{\tau},\tau,c^{o})\bigr\rVert_2^2,
\end{equation}
sharing a common noise schedule. Boxes and trajectories are denoised using a lightweight 1D U-Net~\cite{ronneberger2015u}, while object shapes are synthesised with a point-based U-Net~\cite{zheng2024point}. Different from LOGen~\cite{kirby2024logen}, we match only the LiDAR sampling distribution, not the exact foreground points, which eliminates the heavy DiT cost~\cite{peebles2023scalable} yet still delivers plausible inputs for later refinement. More details about the denoiser are given in the Appendix.

\clearpage\clearpage
\subsection{Layout2Scene: Controlled LiDAR Generation}
\label{sec:static_pointcloud_generation}
\begin{wrapfigure}{r}{0.56\textwidth}
    \begin{minipage}{\linewidth}
        \centering
        \includegraphics[width=\linewidth]{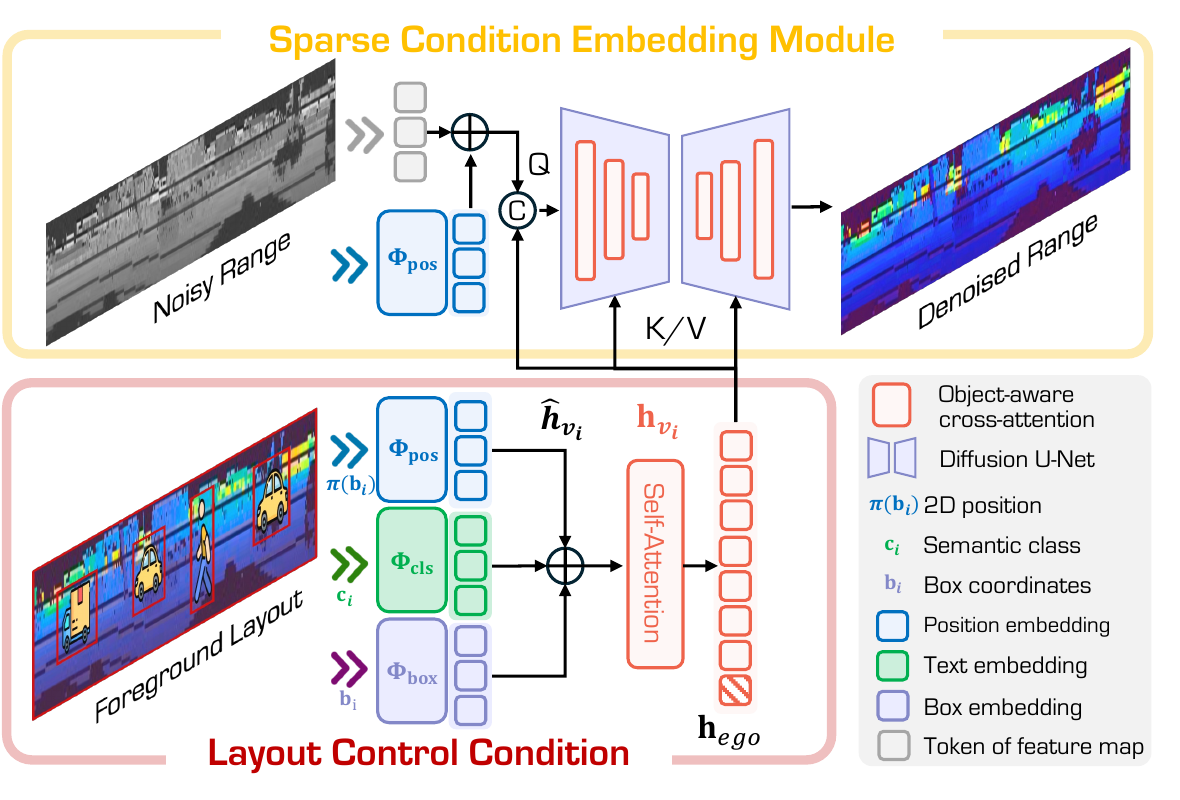}
        \vspace{-0.5cm}
        \caption{Structures of our range-image LiDAR diffusion model.}
        \label{fig:condition_embedding}
    \end{minipage}
\end{wrapfigure}
\textbf{LiDARCrafter} ensures generation fidelity by using a unified range-image diffusion backbone that generates LiDAR point clouds end to end. Given the scene graph \(\mathcal{G}\) and the decoded layout \(\mathcal{O}_i\), the network denoises Gaussian noise into the clean range frame \(\mathbf{I}^0\), thereby bootstrapping the LiDAR sequence \(\mathcal{P}=\{\mathbf{P}^t\}_{t=0}^{T}\) while following the pose, trajectory, and coarse shape of each object. The range-view representation preserves native LiDAR geometry while remaining convolution-friendly~\cite{nakashima2024lidar,kong2023rangeformer,xu2025frnet,kong2025multi}.

\noindent\textbf{Sparse Object Conditioning.} Directly projecting all foreground points into the range image, as in OLiDM~\cite{yan2025olidm}, inadequately represents small or distant objects (\emph{e.g.}, a car at $15$m may occupy only a few dozen pixels). To address this, we condition the model on sparse object representations that encode semantics, pose, and coarse shape, thereby enabling the model to hallucinate fine structure, as shown in Figure~\ref{fig:condition_embedding}.
For each node, we aggregate its features  
\begin{equation}
    \hat{\mathbf{h}}_{v_i} = \Phi_{\text{pos}}\bigl(\pi(\mathbf{b}_i)\bigr) + \Phi_{\text{cls}}(c_i) + \Phi_{\text{box}}(\mathbf{b}_i),
\end{equation}
where \(\pi(\mathbf{b}_i)\) is the 3D box projected to image coordinates, \(\Phi_{\text{pos}}\) is a positional embedder, and \(\Phi_{\text{cls}},\Phi_{\text{box}}\) are learned MLPs. A lightweight self-attention layer diffuses contextual cues across tokens~\cite{vaswani2017attention}, producing the refined vector \({\mathbf{h}}_{v_i}\).
The ego token is further compressed by an MLP to form a scene-level vector \(\mathbf{h}_{\text{ego}}\).

During denoising step \(\tau\),  the noisy range map $\mathbf{d}_{\tau}$ is concatenated with a sparse conditioning map \(\mathbf{I}_{\text{cond}}\) as model input, which is obtained by projecting all layout points \(\{\mathbf{p}_i\}_{i=0}^{M}\) onto the image plane. The global context is formed by summing the scene-level vector, a time embedding, and a CLIP embedding of the ego state: 
\begin{equation}
    \mathbf{h}_{\text{cond}}
    \;=\;
    \mathbf{h}_{\text{ego}}
    + \Phi_{\text{time}}(\tau)
    + \text{CLIP}(s_0).
\end{equation}
A transformer-based U-Net then predicts the clean signal, progressively sharpening geometry and semantics.

\vspace{0.5mm}
\noindent\textbf{Layout-Driven Scene Editing.} As each frame is anchored by an explicit layout, we can edit objects without disturbing the static background, which is crucial for testing planners. After the original scene \(\mathbf{d}_0^{\text{orig}}\) is synthesized, a user may alter the layout tuple.  We then rerun the reverse diffusion, preserving pixels whose 2D projections remain unchanged following~\cite{lugmayr2022repaint}.  At each denoising step:
\begin{equation}
\mathbf{d}_{\tau-1} \;=\;
(1-\mathbf{m})\odot\tilde{\mathbf{d}}_{\tau-1}
\;+\;
\mathbf{m}\odot\hat{\mathbf{d}}_{\tau-1},
\end{equation}
where \(\hat{\mathbf{d}}_{\tau-1}\) is the freshly denoised sample, \(\tilde{\mathbf{d}}_{\tau-1}\sim\mathcal{N}\!\bigl(\sqrt{\bar{\alpha}}\, \mathbf{d}_0^{\mathrm{orig}},\, (1-\bar{\alpha})\mathbb{I}\bigr)\) is a Gaussian-perturbed copy of the original scene, and the binary mask \(\mathbf{m}\) marks pixels affected by the edited boxes. The blend locks untouched regions and resynthesizes only the modified objects, delivering instant, artifact-free edits for closed-loop simulation.

\begin{figure}[t]
    \centering
    \includegraphics[width=.8\linewidth]{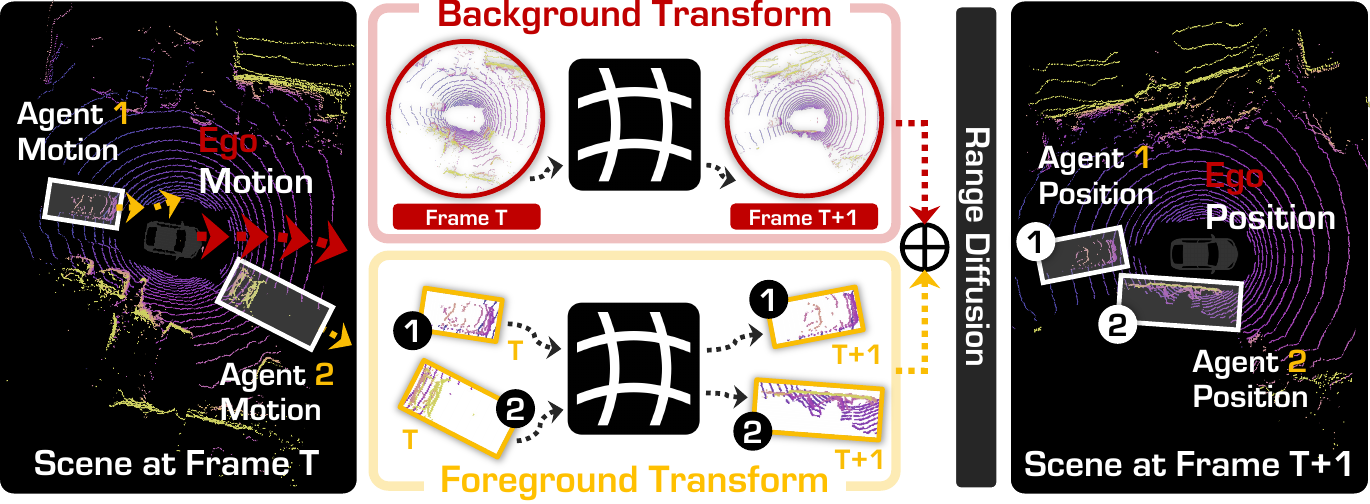}
    \vspace{-0.2cm}
    \caption{Details of the foreground and background warp.}
    \label{fig:autoregress}
\end{figure}

\subsection{Scene2Seq: Autoregressive LiDAR Synthesis}
\label{sec:4D_pointcloud_generation}
A core innovation of \textbf{LiDARCrafter} is its ability to generate the LiDAR stream autoregressively. In RGB video, textures and lighting change constantly, whereas a LiDAR sweep sees a mostly static environment, with only the ego vehicle and annotated objects moving. We exploit this stability by warping previously observed points to create a strong prior, as shown in Figure~\ref{fig:autoregress}. Concretely, we back-project the first range image \(\mathbf{I}^0\) to a point cloud \(\mathbf{P}^0\), then split it with the layout boxes into background \(\mathbf{B}^{0}\) and foreground sets
\(\{\mathbf{F}^{0}_{i}\}_{i=1}^{M}\). In later frames, we warp \(\mathbf{B}^{0}\) with the ego pose, and update each \(\mathbf{F}^{0}_{i}\) by its own motion prior, providing a strong drift-free geometric prior for the diffusion model at every denoising step.

\vspace{0.5mm}
\noindent\textbf{Static-Scene Warp.}
We update the background points with the ego pose.  Taking frame $0$ as the world origin, the ego translation at step \(t\) is \(\mathbf{u}_{0}^{t}=[\Delta x_{0}^{t},\Delta y_{0}^{t},z_{0}]^{\top}\), with $z_{0}$ is the fixed sensor height,  and its incremental yaw is
\begin{equation}
\label{eq:yaw_angle}
\psi^t_{0} = \mathrm{atan2}(\Delta y^t_{0} - \Delta y^{t-1}_{0},\,\Delta x^t_{0} - \Delta x^{t-1}_{0}).
\end{equation}
We form the homogeneous ego pose matrix $\mathbf{G}_{0}^{t}\in \text{SE}(3)$ with the rotation matrix $\mathbf{R}_{z}(\psi_{0}^{t})$ and translation $\mathbf{u}_{0}^{t}$, and compute the relative motion \(\Delta\mathbf{G}_{0}^{t}=\mathbf{G}_{0}^{t}(\mathbf{G}_{0}^{t-1})^{-1}\), then we propagate the static cloud via \(\mathbf{B}^{ t}=\Delta\mathbf{G}_{0}^{t}\mathbf{B}^{t-1}\).

\vspace{0.5mm}
\noindent\textbf{Dynamic-Object Warp.}
For each object \(i\), we first shift its box center by its own cumulative planar offsets \((\Delta x_i^t,\Delta y_i^t)\), giving the world-frame position \(\mathbf{u}_i^t=[x_i+\Delta x_i^t,\;y_i+\Delta y_i^t,\;z_i]^{\top}\). Its heading change \(\psi_i^t\) is obtained exactly as in Equation~\ref{eq:yaw_angle}. To express the box in the current ego frame, we apply the inverse ego transform: first translate by \(-\mathbf{u}_0^{t}\) and then rotate by \(-\psi_0^{t}\). The same rigid transform maps the stored foreground points \(\mathbf{F}_i^{0}\) to \(\mathbf{F}_i^{t}\). These warped foreground object points, combined with the updated background points, supply a strong geometric prior for the later timestep.

\vspace{0.5mm}
\noindent\textbf{Autoregressive Generation.}
At every timestep \(t\!>\!0\) we build a condition range map by projecting and combining,
\begin{equation}
    I^t_{\text{cond}}
= \Pi\bigl(\mathbf{B}^{0\to t}\cup \mathbf{B}^{t-1\to t}\cup \{\mathbf{F}^{t-1\to t}_i\}_{i=1}^M\bigr),
\end{equation}
where \(\Pi(\cdot)\) denotes spherical projection, and the superscript indicates the warp between two timestamps. Including the first frame background warp \(\mathbf{B}^{0\rightarrow t}\) eliminates accumulated drift. We concatenate \(I_{\text{cond}}^{t}\) with the noisy sample and feed it into the diffusion backbone to generate the next range image, iterating until the whole sequence is synthesized.

\subsection{EvalSuite: Temporal \& Semantic Scoring}
\label{sec:evaluation}
Existing LiDAR generation metrics like FRD judge only static realism. They ignore object semantics, layout validity, and motion coherence, which are essential for a controllable 4D world model. Our EvalSuite adds targeted scores for each facet. \textbf{Object metrics} (FDC, CDA, CFCA, CFSC) verify that generated foreground clouds carry the right labels, box geometry, and detector confidence. \textbf{Layout metrics} (SCR, MSCR, BCR, TCR) measure spatial and trajectory consistency while penalizing box or path collisions. \textbf{Temporal metrics} (TTCE, CTC) track frame-to-frame transform accuracy and sequence smoothness. Together, these metrics give a complete, 4D-aware assessment. More details are given in the appendix.

%% file: sections/4_experiments.tex
\section{Experiments}
\label{sec:experiments}

\begin{figure*}[t]
    \centering
    \includegraphics[width=\textwidth]{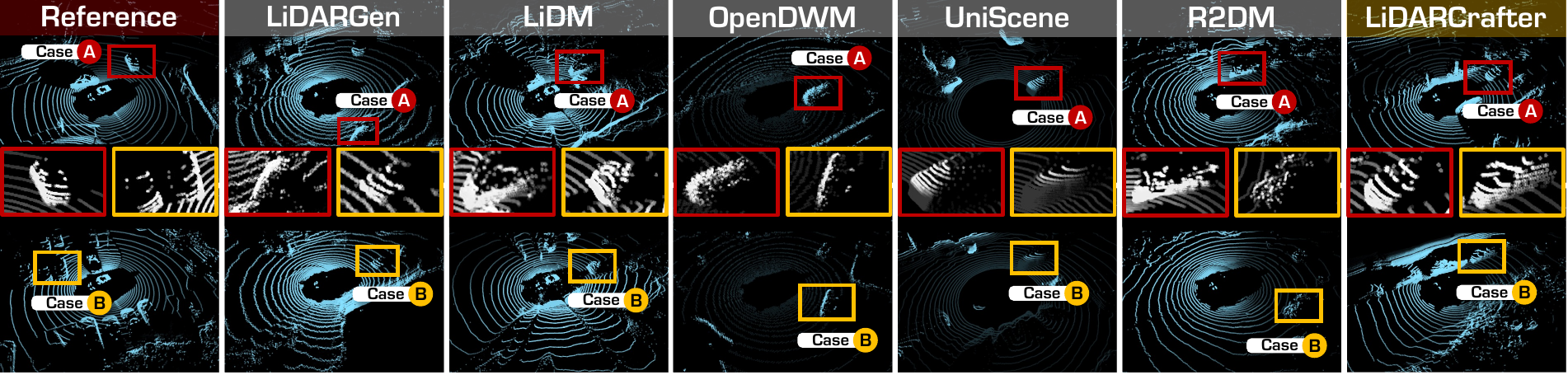}
    \vspace{-0.6cm}
    \caption{
    Single-frame LiDAR point cloud generation results. LiDARCrafter produces the pattern closest to the ground truth, with notably superior foreground quality compared to other methods. Best viewed at high resolution.
    }
    \label{fig:scene_comparison}
\vspace{-0.2cm}
\end{figure*}

\subsection{Experimental Settings}
\label{sec:exp_settings}

We benchmark \textbf{LiDARCrafter} on the widely used nuScenes dataset~\cite{caesar2020nuscenes}, whose 32-beam LiDAR sweeps, 3D boxes, instance IDs, and HD maps provide rich conditioning cues. Evaluation combines classic static scores (FRD, FPD, JSD, MMD) with our object-, layout-, and motion-centric metrics from Section~\ref{sec:evaluation}. Training uses six NVIDIA A40 GPUs, the three-branch layout diffuser runs $1M$ steps at batch $64$, and the range-image model ($32\times1024$) runs $500K$ steps at batch $32$. All diffusers adopt $1024$ denoise steps for training and $256$ for sampling. More experimental settings appear in the appendix.

\subsection{Scene-Level LiDAR Generation}
\label{sec:comparative}
This section evaluates the overall quality of generated LiDAR scenes, focusing on both \emph{whole-scene fidelity} and the accuracy of synthesized \emph{foreground objects}.

\begin{table}[htbp]
\centering
\begin{minipage}{0.48\linewidth}
  \centering
\input{tables/single_frame_generation_results}
\end{minipage}
\hfill
\begin{minipage}{0.48\linewidth}
  \centering
  \input{tables/objects_detection_confidence_score.tex}
\end{minipage}
\end{table}

\begin{table}[htbp]
\centering
\begin{minipage}{0.48\linewidth}
  \centering
\input{tables/object_detection}
\end{minipage}
\hfill
\begin{minipage}{0.48\linewidth}
  \centering
  \input{tables/obj_generation_metric}
\end{minipage}
\end{table}

\begin{table}[htbp]
\centering
\begin{minipage}{0.48\linewidth}
  \centering
\input{tables/obj_cls_reg}
\end{minipage}
\hfill
\begin{minipage}{0.48\linewidth}
  \centering
  \input{tables/seq_results}
\end{minipage}
\end{table}

\begin{table}[htbp]
\centering
\begin{minipage}{0.48\linewidth}
  \centering
\input{tables/ablation_condition}
\end{minipage}
\hfill
\begin{minipage}{0.48\linewidth}
  \centering
  \input{tables/ablation_autoregressive}
\end{minipage}
\end{table}

\noindent\textbf{Whole-Scene Fidelity.}
Table~\ref{tab:scene_point_generation} compares \ours{} against recent LiDAR generation baselines in terms of whole-scene fidelity. Our method outperforms all existing methods, achieving the lowest FRD ($194.37$, a~$20\% $improvement over R2DM) and FPD ($8.64$), indicating more accurate reconstruction of both 2D range structures and 3D point distributions. As shown in Figure~\ref{fig:scene_comparison}, our method produces scans that closely resemble the ground truth, with well-preserved foreground structures, while other methods suffer from background noise or blurring.

\noindent\textbf{Foreground Object Accuracy.}
To evaluate the fidelity of synthesized foregrounds, we apply a pre-trained VoxelRCNN detector~\cite{deng2021voxel} to the generated scenes. We report Foreground Detection Confidence (FDC), which reflects the detector's confidence (\emph{cf.}~Table~\ref{tab:fdc}), and Conditioned Detection Accuracy (CDA), which measures the average precision (AP) of detected boxes (\emph{cf.}~Table~\ref{tab:cda}). LiDARCrafter achieves the highest FDC scores across most categories, indicating superior object fidelity, and also attains the best detection AP, demonstrating stronger alignment between synthesized structures and the given conditions.

\subsection{Object-Level LiDAR Generation}
This section evaluates the quality of individual object generation, focusing on both \emph{fidelity} and \emph{semantic and geometric consistency} under box-level conditioning.

\noindent\textbf{Object-Wise Fidelity.}  
To assess instance-level fidelity, we extract $2{,}000$ \emph{Car} objects from each method and compute object-level metrics (Table~\ref{tab:object_metrics}). LiDARCrafter achieves the lowest FPD ($1.03$) and MMD ($5.48$), significantly outperforming OpenDWM and demonstrating better reconstruction of fine-grained geometry.

\noindent\textbf{Semantic and Geometric Consistency.}
To further evaluate object quality under conditioning, we introduce two metrics: CFCA for semantic fidelity and CFSC for geometric consistency (Table~\ref{tab:cfca_cfsc}). For \emph{semantic fidelity}, we apply a PointMLP~\cite{ma2022rethinking} classifier trained on real data to classify generated instances, yielding a CFCA score of $73.48\%$ for LiDARCrafter, indicating strong alignment with real-world categories. For \emph{geometric consistency}, we use a conditional variational autoencoder to regress bounding boxes from generated point clouds, and compute the mean IoU with ground truth. LiDARCrafter achieves the highest IoU across all point count settings, demonstrating superior adherence to geometric constraints.

\begin{figure*}[t]
    \centering
    \includegraphics[width=\textwidth]{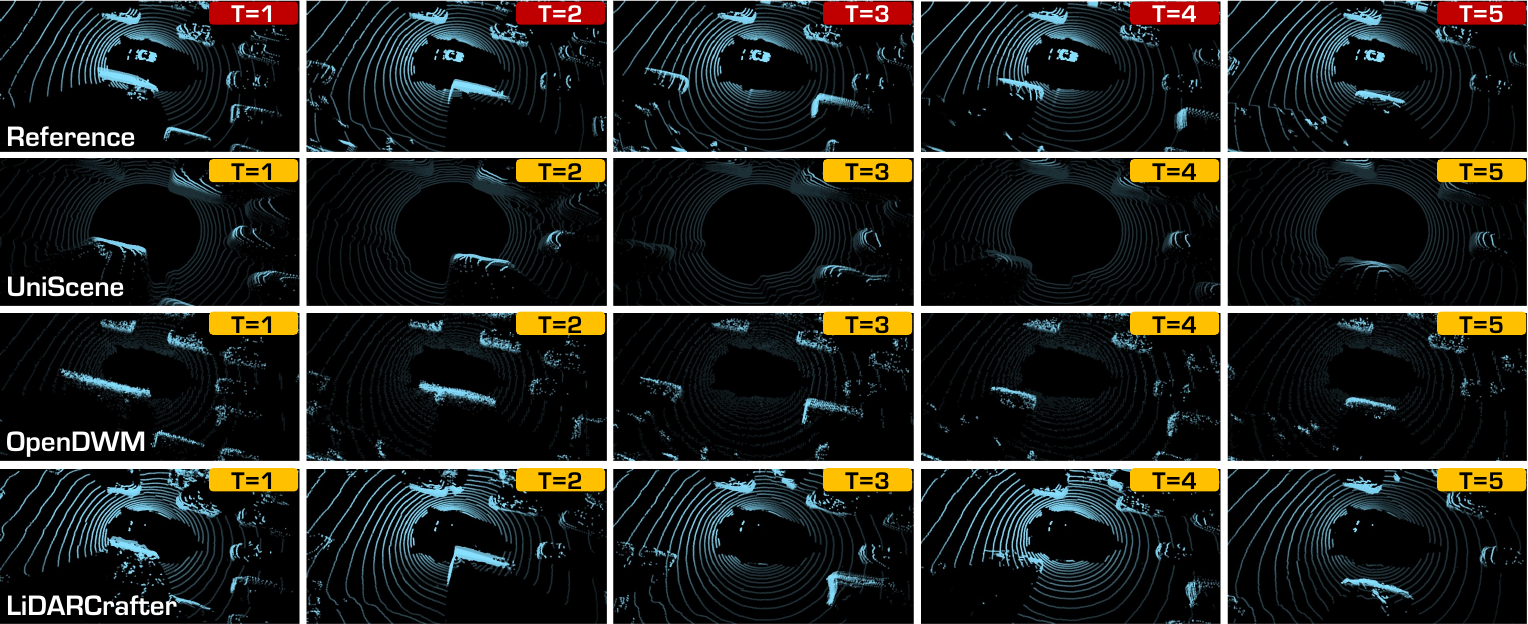}
    \vspace{-0.6cm}
    \caption{
    Sequence point cloud generation results. LiDARCrafter maintains temporal consistency while producing patterns closest to the ground truth. Frames are arranged in temporal order from left to right. Best viewed at high resolution. 
    }
    \label{fig:sequence_comparison}
\end{figure*}

\subsection{Autoregressive 4D LiDAR Generation}
\noindent\textbf{Temporal Consistency.}
We evaluate temporal consistency in 4D LiDAR generation in~Table~\ref{tab:sequence_coherence}. TTCE measures the error between the predicted and ground-truth transformation matrices obtained via point cloud registration, while CTC computes the Chamfer Distance between consecutive frames. Our approach achieves the lowest TTCE scores across both frame intervals and maintains competitive CTC performance at all intervals, demonstrating strong temporal coherence. Qualitative comparisons in Figure~\ref{fig:sequence_comparison} further show that LiDARCrafter produces sequences with consistent structure and fine geometric detail, whereas other methods often suffer from degraded fidelity over time.

\subsection{Ablation Study}
\label{sec:ablation}
We conduct ablations on \textbf{foreground generation} (necessity and conditioning) and \textbf{4D consistency} (generation paradigm and historical priors) to validate key designs.

\begin{wrapfigure}{r}{0.6\textwidth}
\begin{minipage}{\linewidth}
    \centering
    \includegraphics[width=\linewidth]{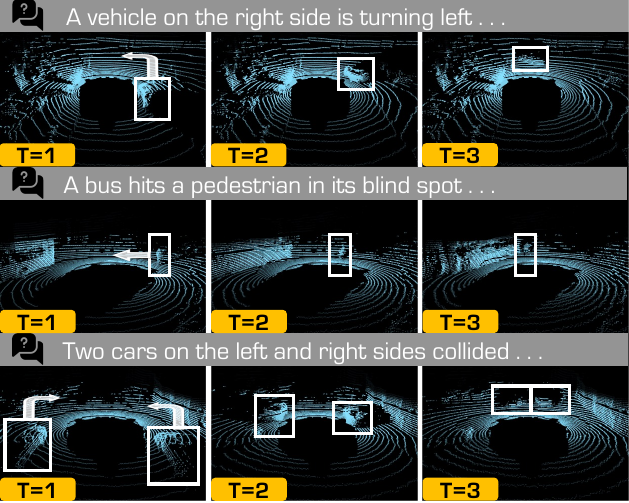}
    \vspace{-0.6cm}
    \caption{
    \textbf{Diverse corner cases} generated by LiDARCrafter with object-centric controllability. Best viewed at high resolution. Frames are arranged sequentially from left to right.
    }
    \label{fig:corner_case}
\end{minipage}
\end{wrapfigure}

\noindent\textbf{Ablation on the necessity of foreground generation.}
Table~\ref{tab:ablation_condition_embedding} shows that introducing a 2D foreground mask projected from 3D boxes \emph{(No.2)} notably improves scene generation, particularly for foreground objects. Further incorporating the foreground generation branch \emph{(No.3)} that produces fine-grained object masks leads to a lower FPD, showing the benefit of detailed geometric and depth supervision.

\noindent\textbf{Ablation on foreground conditioning mechanism.}
Foreground objects are inherently sparse, often occupying only a few pixels, making dense mask-only conditioning insufficient. As shown in Table~\ref{tab:ablation_condition_embedding}, our proposed \textbf{sparse conditioning modules} are crucial: embedding 2D box features alone \emph{(No.4)} reduces FRD, while adding semantic and geometric attributes \emph{(No.5 and No.6)} yields the best FPD and further improves FRD. These results underscore the benefits of richer, object-centric conditioning.

\noindent\textbf{Ablation on generation paradigm in 4D generation.}
Unlike RGB videos, where appearance varies due to lighting and texture changes, LiDAR sequences capture largely static environments, with dynamics introduced only by ego-motion and moving agents. We exploit this stability by warping previously observed points using ego and object trajectories, providing strong priors for \textbf{autoregressive generation}. As shown in Table~\ref{tab:ablation_ar}, our inpainting-based autoregressive framework \emph{(No.2)} outperforms the end-to-end baseline \emph{(No.1)} on temporal metrics, demonstrating that the autoregressive design naturally aligns with the relatively static nature and limited temporal variation of LiDAR sequences.

\noindent\textbf{Ablation on historical conditioning in 4D generation.}
Table~\ref{tab:ablation_ar} investigates the impact of different historical priors on 4D LiDAR generation. Using both depth and intensity features as conditioning inputs \emph{(No.3)} significantly improves performance over the baseline without historical guidance \emph{(No.2)}. Notably, excluding the depth prior \emph{(No.4)} leads to substantial error accumulation (FRD increases by 109.88 compared to \emph{No.3}), while using depth alone \emph{(No.5)} achieves the best FRD. These results indicate that depth cues are more reliable and crucial for maintaining temporal consistency, whereas intensity features are harder to model effectively.

\subsection{Applications}
\label{sec:Applications}
Leveraging its object-centric generation capability, \ours{} enables the synthesis of rare and diverse corner-case scenarios, which are valuable for data augmentation and for evaluating the robustness of downstream algorithms. As illustrated in Figure~\ref{fig:corner_case}, our method can generate challenging situations such as a vehicle cutting into the driving lane, a pedestrian emerging in a bus blind spot, a collision between two adjacent cars, and an overtaking maneuver. These results demonstrate that our method not only produces the desired scenarios on demand, but also maintains strong temporal coherence throughout the sequence. Additional qualitative examples are provided in the appendix.

%% file: tables/single_frame_generation_results.tex
    \centering
    \caption{Evaluations of \textbf{scene-level fidelity} for LiDAR generation on the \textit{nuScenes} dataset. MMD values are reported in $10^{-4}$ and JSD in $10^{-2}$. Lower is better for all metrics.} 
    \vspace{-0.3cm}
    \resizebox{\linewidth}{!}{
    \begin{tabular}{c|r|r|cc|cc|cc}
    \toprule
    \multirow{2}{*}{\textbf{\#}} & \multirow{2}{*}{\textbf{Method}} & \multirow{2}{*}{\textbf{Venue}} & \multicolumn{2}{c|}{\textbf{Range}} & \multicolumn{2}{c|}{\textbf{Points}} & \multicolumn{2}{c}{\textbf{BEV}} 
    \\
    & & & \textbf{FRD}$\downarrow$ & \textbf{MMD}$\downarrow$ & \textbf{FPD}$\downarrow$ & \textbf{MMD}$\downarrow$ & \textbf{JSD}$\downarrow$ & \textbf{MMD}$\downarrow$ 
    \\
    \midrule\midrule
    \multirow{3.5}{*}{\rotatebox{90}{\textbf{Voxel}}}
    &UniScene & CVPR'25 &      –   &   –  & $976.47$ & $29.06$ & $31.55$ & $13.61$ \\
    &OpenDWM       & CVPR'25  &      –   &   –  & $714.19$ & $21.95$ & $20.17$ & $5.61$ \\
    &OpenDWM-DiT   & CVPR'25  &      –   &   –  & $381.91$ & $12.46$ & $19.90$ & $5.73$ \\
    \midrule
    \multirow{5.5}{*}{\rotatebox{90}{\textbf{Range}}}    
    &LiDARGen & ECCV'22  & $759.65$ & $1.71$ & $159.35$ & $35.52$ & $5.74$ & $2.39$ \\
    &LiDM          & CVPR'24  &  $495.54$ & $0.18$ & $210.20$ &  $8.45$ & $5.86$ & $0.73$ \\
    &RangeLDM      & ECCV'24  &      –   &   –  &     –   &    –   & $5.47$ & $1.92$ \\
    &R2DM          & ICRA'24  &  $243.35$ & $1.40$ &  $33.97$ &  $1.62$ & $3.51$ & $0.71$ \\
    \cmidrule{2-9}
    &\textbf{LiDARCrafter} & \textbf{Ours} & \cellcolor{crafter!15}$\mathbf{194.37}$ & \cellcolor{crafter!15}$\mathbf{0.08}$  & \cellcolor{crafter!15}$\mathbf{8.64}$ & \cellcolor{crafter!15}$\mathbf{0.90}$ & \cellcolor{crafter!15}$\mathbf{3.11}$ & \cellcolor{crafter!15}$\mathbf{0.42}$ 
    \\
    \bottomrule
    \end{tabular}}
    \label{tab:scene_point_generation}

%% file: tables/objects_detection_confidence_score.tex
    \centering
    \caption{
    Comparison of \textbf{foreground object quality} using FDC ($\uparrow$), which reflects detector confidence on generated scenes. \#Box is the average number of boxes per frame.
    }
    \vspace{-0.3cm}
    \resizebox{\linewidth}{!}{
    \begin{tabular}{c|r|r|cccc|c}
    \toprule
    \textbf{\#} & \textbf{Method} & \textbf{Venue} & \textbf{Car}$\uparrow$ & \textbf{Ped}$\uparrow$ & \textbf{Truck}$\uparrow$ & \textbf{Bus}$\uparrow$ 
      & \textbf{\#Box} \\
    \midrule\midrule
    \multirow{3.1}{*}{\rotatebox{90}{\small \textbf{Uncond.}}}
    &LiDARGen      & ECCV'22  
      & $$0.57$$ & $$0.29$$ & $$0.42$$ & $$0.38$$ 
      & $$0.364$$ \\
    &LiDM          & CVPR'24  
      & $$0.65$$ & $$0.22$$ & $$0.45$$ & $$0.31$$ 
      & $$0.28$$  \\
    &R2DM          & ICRA'24  
      & $$0.54$$ & $$0.29$$ & $$0.39$$ & $$0.35$$ 
      & $$0.53$$  \\
    \midrule
    \multirow{4.5}{*}{\rotatebox{90}{\textbf{Cond.}}}
    &UniScene      & CVPR'25  
      & $$0.53$$ & $$0.28$$ & $$0.35$$ & $$0.25$$ 
      & $$0.98$$  \\
    &OpenDWM       & CVPR'25  
      & $$0.74$$ & $$0.30$$ & $$0.51$$ & $$0.44$$ 
      & $$0.54$$  \\
    &OpenDWM-DiT   & CVPR'25  
      & $$0.78$$ & $$0.32$$ & $\mathbf{0.56}$ & $$0.51$$ 
      & $$0.64$$  \\
    \cmidrule{2-8}
    &\textbf{LiDARCrafter} & \textbf{Ours} & \cellcolor{crafter!15}$\mathbf{0.83}$ & \cellcolor{crafter!15}$\mathbf{0.34}$ & \cellcolor{crafter!15}$0.55$ & \cellcolor{crafter!15}$\mathbf{0.54}$ & \cellcolor{crafter!15}$\mathbf{1.84}$ 
    \\
    \bottomrule
    \end{tabular}}
    \label{tab:fdc}

%% file: tables/object_detection.tex
    \centering
    \caption{
    Comparisons of \textbf{foreground object perception} accuracy using the CDA ($\uparrow$) metric, which measures 3D detection average precision (AP) on generated LiDAR scenes.
    }
    \vspace{-0.3cm}
    \resizebox{\linewidth}{!}{
    \begin{tabular}{r|r|cccc}
    \toprule
    \textbf{Method}       & \textbf{Venue} 
      & $\textbf{AP}^\mathrm{R11}_\mathrm{BEV}$
      & $\textbf{AP}^\mathrm{R11}_\mathrm{3D}$
      & $\textbf{AP}^\mathrm{R40}_\mathrm{BEV}$
      & $\textbf{AP}^\mathrm{R40}_\mathrm{3D}$ 
      \\\midrule\midrule
    UniScene              & CVPR'25 
      & $$0.19$$   & $$0$$   
      & $$0.02$$  & $$0$$    
      \\
    OpenDWM               & CVPR'25 
      & $$17.07$$  & $$9.09$$
      & $$11.84$$ & $$1.03$$ 
      \\
    OpenDWM-DiT           & CVPR'25 
      & $$16.37$$  & $$11.27$$
      & $$10.62$$ & $$1.89$$ 
    \\\midrule
    \textbf{LiDARCrafter} & \textbf{Ours} & \cellcolor{crafter!15}$\mathbf{23.21}$ & \cellcolor{crafter!15}$\mathbf{15.24}$ & \cellcolor{crafter!15}$\mathbf{18.27}$ & \cellcolor{crafter!15}$\mathbf{8.26}$ 
    \\
    \bottomrule
    \end{tabular}}
    \label{tab:cda}

%% file: tables/obj_generation_metric.tex
    \centering
    \caption{
    Evaluation of \textbf{object-level fidelity} for LiDAR generation. MMD is reported in $10^{-4}$, and JSD in $10^{-2}$.
    }
    \vspace{-0.3cm}
    \resizebox{\linewidth}{!}{
    \begin{tabular}{c|r|r|cccc}
    \toprule
    \textbf{\#}& \textbf{Method}       & \textbf{Venue} 
      & \textbf{FPD}$\downarrow$ 
      & \textbf{P-MMD}$\downarrow$ 
      & \textbf{JSD}$\downarrow$ 
      & \textbf{MMD}$\downarrow$ 
    \\
    \midrule\midrule
    \multirow{3.1}{*}{\rotatebox{90}{\small \textbf{Uncond.}}}
    &LiDARGen      & ECCV'22 
      & $$1.39$$  & $$0.15$$ 
      & $$0.20$$  & $$16.22$$ 
    \\
    &LiDM          & CVPR'24 
      & $$1.41$$  & $$0.15$$ 
      & $$0.19$$  & $$13.49$$ 
    \\
    &R2DM          & ICRA'24 
      & $$1.40$$  & $$0.15$$ 
      & $$0.17$$  & $$12.76$$ 
    \\
    \midrule
    \multirow{4.5}{*}{\rotatebox{90}{\textbf{Cond.}}}
    &UniScene      & CVPR'25 
      & $$1.19$$  & $$0.18$$ 
      & $$0.23$$  & $$16.65$$ 
    \\
    &OpenDWM       & CVPR'25 
      & $$1.49$$  & $$0.19$$ 
      & $$0.16$$  & $$9.11$$  
    \\
    &OpenDWM-DiT   & CVPR'25 
      & $$1.48$$  & $$0.18$$ 
      & $\mathbf{0.15}$  & $$9.02$$  
    \\
    \cmidrule{2-7}
    &\textbf{LiDARCrafter} & \textbf{Ours} 
    & \cellcolor{crafter!15}$\mathbf{1.03}$ & \cellcolor{crafter!15}$\mathbf{0.13}$ & \cellcolor{crafter!15}$\mathbf{0.15}$ & \cellcolor{crafter!15}$\mathbf{5.48}$ 
    \\
    \bottomrule
    \end{tabular}}
    \label{tab:object_metrics}

%% file: tables/obj_cls_reg.tex
    \centering
    \caption{
    Comparison of \textbf{object generation consistency} using CFCA ($\uparrow$) and CFSC ($\uparrow$). CFCA measures classification accuracy on generated points using a PointMLP trained on real data. CFSC assesses geometric consistency by regressing boxes from generated points and computing IoU. Numbers indicate the point count within each box.}
    \vspace{-0.2cm}
    \resizebox{\linewidth}{!}{
    \begin{tabular}{r|r|c|ccc}
    \toprule
    \multirow{2}{*}{\textbf{Method}} 
      & \multirow{2}{*}{\textbf{Venue}} 
      & \multirow{2}{*}{\textbf{CFCA}$\uparrow$} 
      & \multicolumn{3}{c}{\textbf{CFSC}$\uparrow$} \\
    & & & \textbf{$<150$} & 150--300 & \textbf{$>300$} \\
    \midrule\midrule
    Original            & --       & $$92.49$$ & $$0.50$$ & $$0.61$$ & $$0.72$$ \\
    UniScene       & CVPR'25  & $$34.25$$ & $$0.14$$ & $$0.17$$ & $$0.23$$ \\
    OpenDWM        & CVPR'25  & $$62.35$$ & $$0.17$$ & $$0.21$$ & $$0.26$$ \\
    OpenDWM-DiT    & CVPR'25  & $$70.65$$ & $$0.31$$ & $$0.32$$ & $$0.34$$ \\
    \midrule
    \textbf{LiDARCrafter} & \textbf{Ours} & \cellcolor{crafter!15}$\mathbf{73.45}$ & \cellcolor{crafter!15}$\mathbf{0.35}$ & \cellcolor{crafter!15}$\mathbf{0.36}$ & \cellcolor{crafter!15}$\mathbf{0.42}$ 
    \\
    \bottomrule
    \end{tabular}}
    \label{tab:cfca_cfsc}

%% file: tables/seq_results.tex
    \centering
    \caption{
    Comparison of \textbf{temporal consistency} in 4D LiDAR generation using TTCE ($\downarrow$) and CTC ($\downarrow$). TTCE measures the transformation error from the point cloud registration perspective, indicating the consistency of temporal changes; CTC computes the Chamfer Distance between the adjacent generated LiDAR frames. Numbers indicate the frame intervals.}
    \vspace{-0.2cm}
    \resizebox{\linewidth}{!}{
    \begin{tabular}{r|r|cc|cccc}
    \toprule
    \multirow{2}{*}{\textbf{Method}} & \multirow{2}{*}{\textbf{Venue}} 
      & \multicolumn{2}{c|}{\textbf{TTCE}$\downarrow$} 
      & \multicolumn{4}{c}{\textbf{CTC}$\downarrow$} \\
    & 
      & \textbf{3} & \textbf{4} 
      & \textbf{1} & \textbf{2} & \textbf{3} & \textbf{4} \\
    \midrule\midrule
    UniScene      & CVPR'25  
      & $$2.74$$ & $$3.69$$ 
      & $$0.90$$ & $$1.84$$ & $$3.64$$ & $\mathbf{3.90}$ \\
    OpenDWM       & CVPR'25  
      & $$2.68$$ & $$3.65$$ 
      & $$1.02$$ & $$2.02$$ & $$3.37$$ & $$5.05$$ \\
    OpenDWM-DiT   & CVPR'25  
      & $$2.71$$ & $$3.66$$ 
      & $\mathbf{0.89}$ & $\mathbf{1.79}$ & $$3.06$$ & $$4.64$$ \\
    \midrule
    \textbf{LiDARCrafter} & \textbf{Ours} & \cellcolor{crafter!15}$\mathbf{2.65}$ & \cellcolor{crafter!15}$\mathbf{3.56}$ & \cellcolor{crafter!15}$1.12$ & \cellcolor{crafter!15}$2.38$ & \cellcolor{crafter!15}$\mathbf{3.02}$ & \cellcolor{crafter!15}$4.81$ 
    \\
    \bottomrule
    \end{tabular}}
    \label{tab:sequence_coherence}

%% file: tables/ablation_condition.tex
    \centering
    \caption{
    Ablation on \textbf{foreground conditioning methods} for generation. 2D masks are projected from 3D boxes.
    }
    \vspace{-0.2cm}
    \resizebox{\linewidth}{!}{
    \begin{tabular}{c|l|l|cc|ccc}
    \toprule
    \multirow{2}{*}{\textbf{No.}} 
      & \multirow{2}{*}{\textbf{Type}} 
      & \multirow{2}{*}{\textbf{Variant}} 
      & \multicolumn{2}{c|}{\textbf{Scene}} 
      & \multicolumn{3}{c}{\textbf{Object}} \\
    & & 
      & \textbf{FRD}$\downarrow$ & \textbf{FPD}$\downarrow$ 
      & \textbf{FPD}$\downarrow$ & \textbf{CFCA}$\uparrow$ & \textbf{CFSC}$\uparrow$ \\
    \midrule\midrule
    1 & Baseline       & --            & $$243.35$$ & $$33.97$$ & $$1.40$$ & --       & --       \\
    \midrule
    2 & \multirow{2}{*}{Dense}        
      & $w/$ 2D mask    & $$237.17$$ & $$33.21$$ & $$1.35$$ & $$61.22$$ & $$0.24$$ \\
    3 &  & $w/$ Obj mask& $$217.83$$ & $$24.02$$ & $$1.20$$ & $$64.54$$ & $$0.27$$ \\
    \midrule
    4 & \multirow{3}{*}{Dense+Sparse} 
     & $w/$ $E_{pos}$     & $$205.27$$ & $$15.97$$ & $$1.08$$ & $$72.46$$ & $$0.40$$ \\
    5 &  & $w/$ $ E_{pos} + E_{cls}$  & $\mathbf{193.27}$ & $$10.52$$ & $$1.05$$ & $\mathbf{75.27}$ & $$0.40$$ \\
    6 &  & $w/$ All        & $$194.37$$ & $\mathbf{8.64}$ & $\mathbf{1.03}$ & $$73.45$$ & $\mathbf{0.42}$ \\
    \bottomrule
    \end{tabular}}
    \label{tab:ablation_condition_embedding}

%% file: tables/ablation_autoregressive.tex
    \centering
    \caption{
    Ablation on \textbf{generation paradigm} (E2E vs. AR) and \textbf{historical conditioning} for 4D LiDAR generation.
    }
    \vspace{-0.2cm}
    \resizebox{\linewidth}{!}{
    \begin{tabular}{c|c|cc|cc|cc|cc}
    \toprule
    \multirow{2}{*}{\textbf{No.}} & \multirow{2}{*}{\textbf{Type}} & \multirow{2}{*}{\textbf{Intensity}} & \multirow{2}{*}{\textbf{Depth}} & \multicolumn{2}{c|}{\textbf{TTCE}$\downarrow$} & \multicolumn{2}{c|}{\textbf{CTC}$\downarrow$} & \multicolumn{2}{c}{\textbf{Scene}} 
    \\
    & & & & \textbf{3} & \textbf{4} & \textbf{3} & \textbf{4} & \textbf{FRD}$\downarrow$ & \textbf{FPD}$\downarrow$ 
    \\
    \midrule\midrule
    1 & {E2E} & - & -
      & $$3.21$$ & $$4.36$$ 
      & $$5.68$$ & $$7.41$$ 
      & $$477.21$$ & $$182.36$$ \\
    \midrule
    2 & \multirow{4}{*}{AR}
      & - & -
        & $$3.31$$ & $$4.84$$ 
        & $$4.31$$ & $$6.21$$ 
        & $$311.27$$ & $$90.10$$ \\
    3 &  & \cmark & \cmark
        & $$2.96$$ & $$3.87$$ 
        & $$3.24$$ & $$4.85$$ 
        & $$254.39$$ & $$22.20$$ \\
    4 &  & \cmark & \xmark
        & $$3.21$$ & $$4.21$$ 
        & $$3.42$$ & $$5.19$$ 
        & $$364.27$$ & $$154.21$$ \\
    5 &  & \xmark & \cmark
        & $\mathbf{2.65}$ & $\mathbf{3.56}$ 
        & $\mathbf{3.02}$ & $\mathbf{4.81}$ 
        & $\mathbf{194.37}$ & $\mathbf{8.64}$  \\
    \bottomrule
    \end{tabular}}
    \label{tab:ablation_ar}

%% file: sections/5_conclusion.tex
\section{Conclusion}
\label{sec:conclusion}
We presented \ours{}, a unified framework for controllable 4D LiDAR sequence generation and editing. By leveraging scene graph descriptors, the multi-branch diffusion model, and an autoregressive generation strategy, our approach achieves fine-grained controllability and strong temporal consistency. Experiments on nuScenes demonstrate clear improvements over existing methods in fidelity, coherence, and controllability. Beyond high-quality data synthesis, LiDARCrafter enables the creation of safety-critical scenarios for robust evaluation of downstream autonomous driving systems. Future work will explore multi-modal extensions and further efficiency improvements.

%% file: supp_sections/1_additional_details.tex
\section{Additional Implementation Details}
\label{supp_sec: additional}

In this section, we provide additional implementation details to facilitate reproducibility. Specifically:
\begin{itemize}
    \item \textbf{Notation Summary} (Section~\ref{notation}): We summarize the notations used throughout the main manuscript for clarity.
    \item \textbf{Scene Graph Construction} (Section~\ref{scene-graph}): We explain how scene graphs are built from dataset annotations, which serve as training data for LiDARCrafter.
    \item \textbf{Training Pipeline} (Section~\ref{training}): We detail the data preprocessing steps, loss functions, and hyperparameter settings used for both 4D layout and point cloud generation.
    \item \textbf{Evaluation Suite} (Section~\ref{evaluation}): We clarify the motivation behind each proposed metric and specify the exact computation procedures.
    \item \textbf{Baseline Models} (Section~\ref{baseline}): We briefly describe the baseline methods used for comparisons.
\end{itemize}

\subsection{Summary of Notations}
\label{notation}
For better readability, the notations used in this work have been summarized in Table~\ref{supp_tab: notations}.

\subsection{Scene Graph Construction from Annotations}
\label{scene-graph}
To train LiDARCrafter, we construct an ego-centric scene graph for each LiDAR frame using annotations from the nuScenes dataset. Each scene graph consists of nodes representing the ego vehicle and nearby foreground objects, along with directed edges that capture their spatial and semantic relationships. The construction pipeline is summarized in Algorithm~\ref{alg:scene_graph}, and detailed as follows:

\vspace{0.5mm}
\noindent\textbf{Foreground Object Filtering.}
We begin by filtering foreground objects based on the following criteria:
\begin{itemize}
    \item Objects must belong to one of the predefined categories:
    \emph{car, truck, construction vehicle, bus, trailer, motorcycle, bicycle, pedestrian}.
    \item Objects must contain at least $30$ LiDAR points.
    \item Objects must be within the LiDAR range, within a 3D volume bounded by $[-80, -80, -8, 80, 80, 8]$ meters in the $x$, $y$, and $z$ axes, respectively.
\end{itemize}

\vspace{0.5mm}
\noindent\textbf{Relationship Computation.}
For each pair of filtered foreground objects, we compute their spatial relationships based on their 3D bounding box attributes (location and dimensions). We define the following relationships:
\begin{itemize}
    \item Spatial positions: \{\emph{front, behind, left, right, close by}\}
    \item Relative size: \{\emph{bigger than, smaller than}\}
    \item Relative height: \{\emph{taller than, shorter than}\}
\end{itemize}
These relationships are calculated using geometric comparisons such as Euclidean distances, relative positional coordinates, volumes, and heights.

\vspace{0.5mm}
\noindent\textbf{Ego Vehicle Relationships.}
We additionally compute the relationships between each foreground object and the ego vehicle, which is represented as a node with a bounding box centered at the origin of the LiDAR coordinate system.

\vspace{0.5mm}
\noindent\textbf{Scene Graph Assembly.}
Finally, we assemble the scene graph by combining the filtered foreground objects, their computed relationships, and the ego node into a structured representation. This graph is then stored as an additional attribute within the original dataset annotation structure.

We repeat this process independently for both training and validation splits of the nuScenes dataset and save the resulting structured scene graph annotations for subsequent model training.

\begin{algorithm}[tb]
\caption{Scene Graph Construction Pipeline}
\label{alg:scene_graph}
\begin{algorithmic}[1]
\REQUIRE Dataset annotations $\mathcal{D}=\{D_i\}_{i=1}^{N}$, category list $C$, relationships $R$
\ENSURE Scene graph information $\mathcal{G}=\{G_i\}_{i=1}^{N}$
\STATE Initialize $\mathcal{G}\leftarrow \emptyset$.
\FOR{$D_i$ in $\mathcal{D}$}
    \STATE Filter objects within sensor range and categories in $C$
    \STATE Further filter objects having more than 30 LiDAR points
    \STATE Obtain filtered boxes $B=\{b_1,b_2,\dots,b_m\}$
    \STATE Initialize relationship set $\mathcal{E}\leftarrow\emptyset$
    \FOR{each box pair $(b_s,b_o)$ in $B\times B$, where $s\neq o$}
        \STATE Compute relationships $r\subseteq R$ using geometric attributes
        \STATE Update $\mathcal{E}\leftarrow\mathcal{E}\cup\{(s,r,o)\}$
    \ENDFOR
    \FOR{each box $b_s$ in $B$}
        \STATE Compute relationships $r_{ego}\subseteq R$ with ego node
        \STATE Update $\mathcal{E}\leftarrow\mathcal{E}\cup\{(s,r_{ego},ego)\}$
    \ENDFOR
    \STATE Assemble scene graph $G_i\leftarrow(B,\mathcal{E})$
    \STATE Update $\mathcal{G}\leftarrow\mathcal{G}\cup\{G_i\}$
\ENDFOR
\RETURN $\mathcal{G}$
\end{algorithmic}
\end{algorithm}

\subsection{Training Configurations}
\label{training}
LiDARCrafter adopts a staged training strategy. In this section, we detail the training procedures for the two stages: tri-branch 4D layout generation and range-based point cloud generation.

\subsubsection{4D Layout Generation.}
We independently train three specialized diffusion branches to generate 3D bounding boxes, trajectories, and object point clouds. Each branch utilizes a dedicated data preprocessing pipeline, denoiser architecture, and loss function, which we describe separately below.

\vspace{0.5mm}
\noindent\textbf{$\triangleright${ 3D Bounding Box Generation Branch.}}
For the 3D bounding box generation branch, each training sample consists of a node feature vector derived from the scene graph, denoted as $\mathbf{h}^{(L)}_{v_i}$, along with an encoded representation of the corresponding bounding box. The encoding process is described as follows:
\begin{itemize}
    \item The center coordinates $(x, y, z)$ are normalized to $[0, 1]$ within the predefined spatial bounds of the dataset.
    \item The box size parameters (width $w$, height $h$, length $l$) are transformed using a logarithmic mapping for scale invariance.
    \item The box orientation (yaw angle) is represented using its sine and cosine values for continuity.
\end{itemize}
This process yields an $(N, 8)$-dimensional feature representing the generated boxes for $N$ objects per frame. These features are produced by a diffusion model implemented as a 1D U-Net operating over the sequence of encoded box vectors.

The training objective for the 3D bounding box generation branch consists of two components:

(1) Diffusion Loss:
This loss follows the standard formulation used in diffusion models. It is defined as:
    \begin{equation}
        \mathcal{L}^{\text{diff}}_b=\mathbb{E}_{t,\mathbf{d},\varepsilon}
        \left\lVert\varepsilon-\varepsilon_{\theta}(\mathbf{d}_t,t,c)\right\rVert_2^2\,,
    \end{equation}
where $\mathbf{d}_t$ is the noisy box encoding at timestep $t$, $c$ denotes the conditioning input (\emph{i.e.}, scene graph node features), and $\varepsilon_{\theta}$ is the denoising network.

(2) Box IoU Loss:
To discourage unrealistic overlap between predicted bounding boxes, we introduce an auxiliary IoU-based penalty. Let ${\hat{b}_i}$ denote the predicted bounding boxes for all $i$ objects in a frame, and $\text{IoU}(\hat{b}_i, \hat{b}j)$ represent the IoU between any pair. The IoU loss is computed as:
    \begin{equation}
        \mathcal{L}^{\text{IoU}} = \frac{1}{N_p} \sum_{i\neq j} \max\left(0, \text{IoU}(\hat{b}_i, \hat{b}_j) - \tau \right)\,,
    \end{equation}
where $N_p$ is the number of unique box pairs and $\tau$ is a small threshold (\emph{e.g.}, $0.01$) beyond which overlapping boxes are penalized.

The total loss combines both components:
    \begin{equation}
        \mathcal{L} = \mathcal{L}^{\text{diff}} + \lambda_{\text{IoU}}\mathcal{L}^{\text{IoU}}\,,
    \end{equation}
where $\lambda_{\text{IoU}}$ is a weighting factor (set to $0.01$ in our experiments) controlling the influence of the IoU penalty. This combined objective encourages accurate box generation while promoting physically consistent object layouts.

\vspace{0.5mm}
\noindent\textbf{$\triangleright${ Trajectory Generation Branch.}}
For the trajectory generation branch, the diffusion model is conditioned on each node’s feature vector $\mathbf{h}^{(L)}_{v_i}$ from the scene graph, along with its encoded 3D bounding box. The model predicts a future trajectory for each object node, represented as a sequence of relative displacements in the $x$ and $y$ directions:
    \[
    \boldsymbol{\delta}_i = \{(\Delta x_i^{\,t}, \Delta y_i^{\,t})\}_{t=1}^{T}\,,
    \]
where $T$ is the number of predicted future frames, set to 5 following the UniScene~\cite{li2025uniscene} configuration. Each displacement is computed relative to the object’s position at the initial frame ($t = 0$). A 1D U-Net serves as the denoising network, operating over the displacement sequence for each object node.

The training objective for the trajectory generation branch comprises two components:

(1) Diffusion Loss. This is the standard loss used in diffusion models, defined as:
    \begin{equation}
        \mathcal{L}^{\text{diff}}_{\text{traj}} = \mathbb{E}_{t,\,\mathbf{d},\,\varepsilon}
    \left\lVert \varepsilon - \varepsilon_{\theta}(\mathbf{d}_t, t, c) \right\rVert_2^2\,,
    \end{equation}
where $\mathbf{d}t$ is the noisy trajectory encoding at timestep $t$, $c$ denotes the conditioning input (\emph{i.e.}, node feature and bounding box), and $\varepsilon_{\theta}$ is the trajectory denoising network.

(2) Trajectory IoU Loss. To discourage physically implausible overlaps during object motion, we introduce a trajectory IoU penalty. Let $\{\hat{\boldsymbol{\delta}}_i\}$ denote the predicted trajectories for all objects, and $\{\hat{b}_i^{\,t}\}$ the corresponding bounding boxes at each timestep $t$ (after applying the predicted displacements). The loss is defined as:
    \begin{equation}
        \mathcal{L}^{\text{IoU}}_{\text{traj}} = \frac{1}{N_p T} \sum_{t=1}^{T}\sum_{i \neq j} \max \left(0, \mathrm{IoU}(\hat{b}_i^{\,t}, \hat{b}_j^{\,t}) - \tau \right)\,,
    \end{equation}
where $N_p$ is the number of unique object pairs and $\tau$ is a small threshold to tolerate minor overlaps.

The total training loss is then given by:
    \begin{equation}
        \mathcal{L}_{\text{traj}} = \mathcal{L}^{\text{diff}}_{\text{traj}} + \lambda_{\text{IoU}} \mathcal{L}^{\text{IoU}}_{\text{traj}}\,,
    \end{equation}
where $\lambda_{\text{IoU}}$ is a weighting hyperparameter (set to $0.01$ in our experiments) that balances the diffusion and IoU losses.

\vspace{0.5mm}
\noindent\textbf{$\triangleright${ Object Point Cloud Generation Branch.}}
The training procedure for the object point cloud generation branch involves normalization and a point-based diffusion model, comprising two main components:

(1) Normalization and Conditioning:
To standardize inputs and enhance conditioning, we normalize both the foreground point cloud and the associated 3D bounding box:
\begin{itemize}
    \item Foreground Point Normalization:
    Each object’s points are rotated to align with the canonical orientation of its bounding box (\emph{i.e.}, subtracting the box's yaw). The coordinates are then scaled by the box dimensions to map $x$, $y$, and $z$ into $[-1, 1]$ relative to the box center. LiDAR intensity values are also normalized to $[-1, 1]$.
    \item Bounding Box Encoding:
    Each bounding box is encoded as a fixed-length vector, including the normalized center position, vertical offset, logarithmic box dimensions, and canonicalized yaw (yaw minus the arctangent of the box center).
    \item Input Construction:
    During training, the bounding box encoding serves as the condition input, and the normalized point cloud is the denoising target.
\end{itemize}
This normalization improves input consistency, strengthens the conditioning effect, and supports stable training.

(2) Denoising Architecture and Objective:
The denoiser is a point-based U-Net following~\cite{zheng2024point}, which directly operates on normalized 3D point sets to capture both local and global structure efficiently.

Training uses the standard diffusion loss:
\begin{equation}
\mathcal{L}^{\text{diff}}_{\text{pts}} = \mathbb{E}_{t,\mathbf{d},\varepsilon}
\left\lVert \varepsilon - \varepsilon_{\theta}(\mathbf{d}_t, t, c) \right\rVert_2^2,
\end{equation}
where $\mathbf{d}t$ is the noisy point set at timestep $t$, $c$ is the bounding box encoding, and $\varepsilon_{\theta}$ is the denoising network.

\vspace{0.5mm}
\noindent\textbf{$\triangleright${ Training Hyperparameters.}}
The tri-branch diffusion network is trained for 1,000,000 steps using a batch size of 32 for training and 64 for evaluation, with 16 data loader workers. We use the Adam optimizer with a learning rate of $1 \times 10^{-4}$, $\beta_1 = 0.9$, $\beta_2 = 0.99$, and $\epsilon = 1 \times 10^{-8}$. A linear warm-up is applied over the first 10,000 steps.
We maintain an exponential moving average (EMA) of model weights with a decay rate of 0.995, updated every 10 steps. Training is performed with mixed precision (fp16) for improved efficiency.
For diffusion, we adopt a cosine noise schedule with 1,024 denoising steps during training and 256 during sampling. The model is trained with $\epsilon$-prediction using an $L_2$ loss, and continuous timesteps are sampled throughout training.
All experiments are conducted using a fixed random seed for reproducibility on a server with five NVIDIA A40 GPUs.

\begin{table}[htbp]
\centering
\begin{minipage}{0.48\linewidth}
  \centering
\input{supp_tables/baselines}
\end{minipage}
\hfill
\begin{minipage}{0.48\linewidth}
  \centering
  \input{supp_tables/layout_gen}
\end{minipage}
\end{table}

\subsubsection{Range-based Point Cloud Generation.}
The static LiDAR point cloud is generated using a separate range-based diffusion model. Below, we detail the corresponding preprocessing steps, model architecture, and training objective.

\vspace{0.5mm}
\noindent\textbf{$\triangleright${ Data Preprocessing.}}
For the range-based point cloud generation stage, each LiDAR frame is first projected into a dense 2D range image via spherical projection. Given a set of 3D points ${p_i^t = (x_i^t, y_i^t, z_i^t)}$ at time $t$, each point is mapped to image coordinates $(u_i^t, v_i^t)$ according to:
    \begin{equation}
    \label{eq:spherical_projection}
    \left(\begin{array}{c}
    u_i^t\\
    v_i^t
    \end{array}\right)
    =
    \left(\begin{array}{c}
    \frac{1}{2}\left[1 - {\arctan(y_i^t, x_i^t)}{\pi^{-1}}\right] w\\
    \left[1 - ({\arcsin(z_i^t / r_i^t) + f_{\mathrm{up}}}){f^{-1}}\right] h
    \end{array}\right)\,,
    \end{equation}
where $r_i^t = \lVert p_i^t \rVert_2$ is the radial distance, $f = f_{\mathrm{up}} - f_{\mathrm{down}}$ is the vertical field of view, and $w$, $h$ denote the predefined width and height of the range image.

At each projected pixel $(u, v)$, we construct a multi-channel feature vector that includes: 3D coordinates ($[x_i^t, y_i^t, z_i^t]$), reflectance ($r_i$), depth ($d_i = r_i^t$), and validity mask ($m_i \in {0, 1}$) indicating valid LiDAR returns.
The resulting image is denoted as $I_t \in \mathbb{R}^{C \times H \times W}$, where $C = 2$ includes the depth and intensity channels.

All input channels are normalized to the range $[-1, 1]$ before being passed to the diffusion model. For depth, we first apply logarithmic compression followed by normalization:
    \begin{equation}
    d_i^{\mathrm{norm}} = \frac{\log_2 (d_i + 1)}{\log_2 (d_{\max} + 1)}\,,
    \end{equation}
and then apply a linear rescaling to match the input range:
    \begin{equation}
    x_{\mathrm{input}} = 2 \cdot x_{\mathrm{norm}} - 1\,,
    \end{equation}
where $x_{\mathrm{norm}} \in [0, 1]$ is the normalized input feature (\emph{e.g.}, depth or intensity).

\vspace{0.5mm}
\noindent\textbf{$\triangleright${ Denoiser \& Loss.}}
The denoising network for range-based point cloud generation is implemented as a 2D U-Net operating on the projected range image. To enable controllable, foreground-aware generation, we incorporate a sparse condition embedding module (as described in the main text). This module encodes the foreground object layout and semantic attributes, and injects them into the U-Net via object-aware cross-attention and positional embeddings at selected layers. This design guides the model to focus on regions of interest while maintaining global scene structure.

We adopt the standard diffusion loss, computed as the mean squared error between the predicted and actual noise added to the range image features, conditioned on the sparse layout embedding. Formally:
    \begin{equation}
        \mathcal{L}^{\text{diff}}_{\text{range}} = \mathbb{E}_{t,\mathbf{d},\varepsilon}
        \left\lVert \varepsilon - \varepsilon_{\theta}(\mathbf{d}_t, t, c) \right\rVert_2^2,
    \end{equation}
where $\mathbf{d}_t$ is the noisy range image at timestep $t$, $c$ is the foreground condition embedding, and $\varepsilon_{\theta}$ is the 2D U-Net denoiser.

This objective encourages the model to generate geometrically and semantically coherent LiDAR range images conditioned on user-specified object layouts.

\input{supp_tables/notations}
\input{supp_tables/metrics}

\vspace{0.5mm}
\subsection{Evaluation Protocols}
\label{evaluation}

We adopt a comprehensive set of metrics to evaluate the quality of scene generation across static, dynamic, and layout dimensions:
\begin{itemize}
    \item \textbf{Scene-Level Evaluations}:
    Fréchet Range Distance (FRD), Fréchet Point Cloud Distance (FPD), Jensen–Shannon Divergence (JSD), and Maximum Mean Discrepancy (MMD).

    \item \textbf{Object-Level Evaluations}:
    Foreground Detection Confidence (FDC), Conditioned Detection Accuracy (CDA), Conditioned Foreground Classification Accuracy (CFCA), and Conditional Foreground Spatial Consistency (CFSC).

    \item \textbf{Temporal-Level Evaluations}:
    Temporal Transformation Consistency Error (TTCE) and Chamfer Temporal Consistency (CTC).

    \item  \textbf{Layout Evaluations}:
    Spatial Consistency Rate (SCR), Motion State Consistency Rate (MSCR), Box Collision Rate (BCR), and Trajectory Collision Rate (TCR).
\end{itemize}

Comprehensive definitions and computation procedures for all metrics are provided below.
Table~\ref{tab:metric_summary} offers a concise summary of all evaluation metrics used in this work, including their hierarchical level, abbreviation, motivation, and references for further details.

\subsubsection{Scene-Level Evaluations.}
To evaluate scene-level fidelity, we adopt standard distributional similarity metrics commonly used in LiDAR generative modeling~\cite{zyrianov2022learning}. All metrics are computed over 10,000 generated samples for statistical reliability.

\vspace{0.5mm}
\noindent\textbf{$\triangleright${ Fréchet Range Distance (FRD).}}
\label{metric_frd}
To evaluate generation quality in the range-image domain, we compute the Fréchet Range Distance (FRD) by measuring the distance between the feature distributions of generated and real range-reflectance images. Following~\cite{zyrianov2022learning}, features are extracted using RangeNet-53~\cite{milioto2019rangenet}, a network pretrained for semantic segmentation on the nuScenes dataset~\cite{caesar2020nuscenes}.

Let $\mathcal{F}_g$ and $\mathcal{F}_r$ denote the sets of features from generated and real samples, with means $\mu_g$, $\mu_r$ and covariances $\Sigma_g$, $\Sigma_r$, respectively. The FRD is defined as:
\begin{equation}
    \mathrm{FRD} = \lVert \mu_g - \mu_r \rVert_2^2 + \mathrm{Tr}\left(\Sigma_g + \Sigma_r - 2 (\Sigma_g \Sigma_r)^{1/2}\right)\,,
\label{eq:frd}
\end{equation}

\vspace{0.5mm}
\noindent\textbf{$\triangleright${ Fréchet Point Distance (FPD).}}
\label{metric_fpd}
To assess generation quality in the 3D point cloud domain, we re-project generated range images back to point clouds and extract features using PointNet~\cite{qi2017pointnet}, pretrained for 16-class object classification on ShapeNet~\cite{chang2015shapenet}. Following~\cite{shu20193d}, the Fréchet Point Distance (FPD) is computed as the distance between the feature distributions of generated and real point clouds:
\begin{equation}
    \mathrm{FPD} = \lVert \mu_g^{\mathrm{pc}} - \mu_r^{\mathrm{pc}} \rVert_2^2 + \mathrm{Tr}\left(\Sigma_g^{\mathrm{pc}} + \Sigma_r^{\mathrm{pc}} - 2 (\Sigma_g^{\mathrm{pc}} \Sigma_r^{\mathrm{pc}})^{1/2}\right)\,,
\label{eq:fpd}
\end{equation}
where $\mu^{\mathrm{pc}}$ and $\Sigma^{\mathrm{pc}}$ denote the mean and covariance of the point cloud feature sets from generated and real samples, respectively.

\vspace{0.5mm}
\noindent\textbf{$\triangleright${ Jensen-Shannon Divergence (JSD).}}
\label{metric_jsd}
To evaluate 2D spatial distributions, we project point clouds into bird’s-eye view (BEV) occupancy maps following~\cite{zyrianov2022learning}. For each sample, a 2D histogram over the BEV grid is computed. The Jensen–Shannon Divergence (JSD) is then calculated between the marginal distributions of generated and real BEV histograms:
\begin{align}
    \mathrm{JSD}(P \Vert Q) &= \frac{\mathrm{KL}(P \Vert M)}{2} + \frac{\mathrm{KL}(Q \Vert M)}{2}\,,\\
    M &= \frac{(P + Q)}{2}\,,
\label{eq:jsd}
\end{align}
where $P$ and $Q$ denote the BEV occupancy distributions of the generated and real samples, respectively.

\vspace{0.5mm}
\noindent\textbf{$\triangleright${ Maximum Mean Discrepancy (MMD).}} 
\label{metric_mmd}
MMD shares the same feature with JSD and is calculated as:
\begin{align}
    \mathrm{MMD}(P, Q) = \left\| \mathbb{E}_{P}[\phi(x)] - \mathbb{E}_{Q}[\phi(y)] \right\|_{\mathcal{H}}^2\,,
\label{eq:mmd}
\end{align}
where $P$ and $Q$ are also BEV occupancy distributions, and $\phi$ is a feature mapping in reproducing kernel Hilbert space.

\subsubsection{Object-Level Evaluations.}
To comprehensively assess the quality of generated foreground objects, we employ a set of metrics that evaluate both scene-level detection performance and object-level geometric fidelity.

\vspace{0.5mm}
\noindent\textbf{$\triangleright${ Foreground Detection Confidence (FDC).}} 
\label{metric_fdc}
FDC is a scene-level metric that evaluates how confidently a pretrained detector recognizes generated foreground objects. We first train a 3D object detector on real nuScenes data and then apply it to each generated point cloud frame. For each object category, we compute the mean detection confidence score across all predicted bounding boxes. A higher FDC indicates that the generated objects are more realistic and easier for the detector to recognize.

Formally, let $\mathcal{D}_k$ denote the set of detected bounding boxes of category $k$ in all generated frames, and let $s_i$ be the confidence score of the $i$-th detection. The FDC for category $k$ is defined as:
\begin{equation}
    \mathrm{FDC}_k = \frac{1}{|\mathcal{D}k|} \sum_{i=1}^{|\mathcal{D}_k|} s_i\,,
\label{eq:fdc}
\end{equation}
where $|\mathcal{D}_k|$ is the total number of detections for class $k$.
FDC can be reported per category or averaged across all categories to provide an overall realism score.

\vspace{0.5mm}
\noindent\textbf{$\triangleright${ Conditioned Detection Accuracy (CDA).}}
\label{metric_cda}
CDA evaluates the quality of generated point clouds conditioned on ground-truth 3D bounding boxes. The core idea is to treat the generated data as if it were real sensor input and assess its utility for downstream detection.

Specifically, we apply a VoxelRCNN detector~\cite{deng2021voxel}, pretrained on real nuScenes data, to the generated point clouds. We then compute standard 3D detection metrics (\emph{e.g.}, Average Precision, AP) by comparing the predicted boxes against the provided ground-truth bounding boxes used during conditioning.
A higher CDA score indicates that the generated scenes are both geometrically accurate and semantically meaningful from the perspective of a pretrained downstream detector.

\vspace{0.5mm}
\noindent\textbf{$\triangleright${ Conditioned Foreground Classification Accuracy (CFCA).}}
\label{metric_cfca}
CFCA is an object-level metric designed to assess the semantic fidelity of generated foreground points. For each generated scene, foreground point sets are extracted using either detected or conditioned 3D bounding boxes.
\begin{itemize}
    \item For unconditional generation, we use high-confidence ($>0.4$) predictions from a pretrained detector to crop foreground regions.
    \item For box-conditioned generation, the cropping is based directly on the provided conditioning boxes.
\end{itemize}
A PointMLP classifier~\cite{ma2022rethinking}, pretrained on ground-truth foreground point clouds, is then used to classify each cropped region. The overall classification accuracy (OA) across all generated foregrounds is reported as the CFCA score. A higher CFCA indicates stronger semantic alignment with real-world object categories.

Formally, let ${\mathcal{P}_j}_{j=1}^{N}$ be the set of cropped foreground point clouds and ${c_j}$ their ground-truth labels. Let $f_\phi$ be the pretrained classifier. CFCA is defined as:
\begin{equation}
    \mathrm{CFCA} = \frac{1}{N} \sum_{j=1}^{N} \mathbb{I}(f_\phi(\mathcal{P}_j) = c_j)\,,
\label{eq:cfca}
\end{equation}
where $\mathbb{I}(\cdot)$ is the indicator function.
This metric quantitatively measures the extent to which generated foreground points exhibit category-consistent semantic features.

\begin{figure}[t]
    \centering
    \includegraphics[width=.9\linewidth]{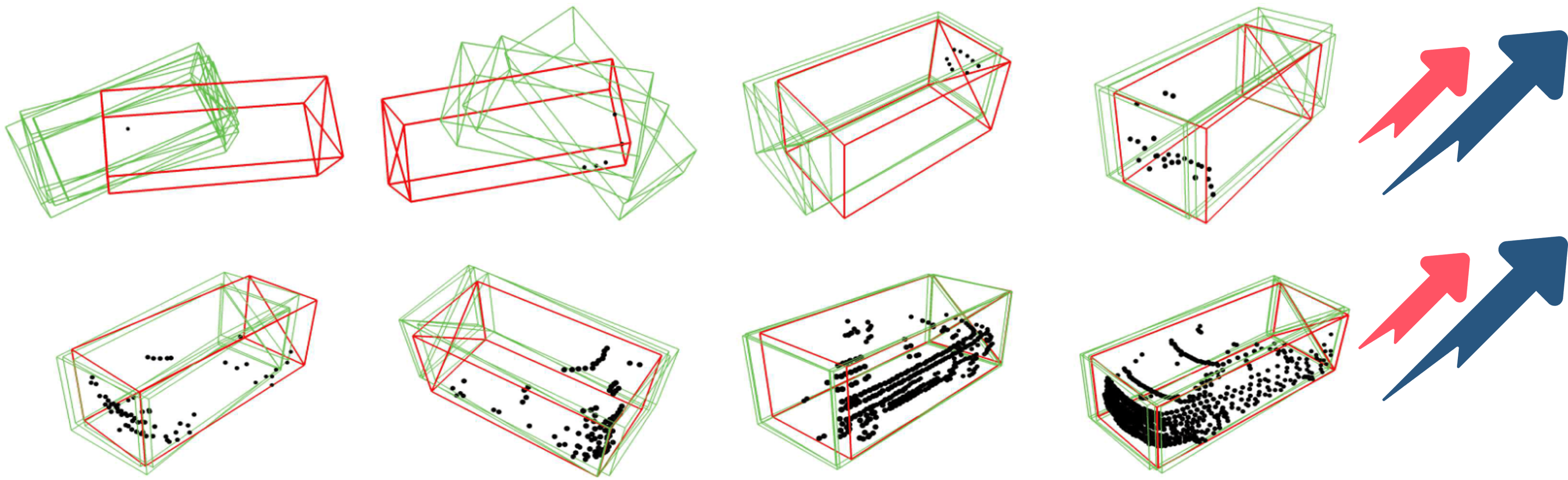}
    \vspace{-0.1cm}
    \caption{
    \textbf{Illustration of Conditional Foreground Spatial Consistency (CFSC)}.
    For each generated foreground object, a CVAE is repeatedly sampled to regress 3D bounding boxes from the synthesized points (green), which are then compared against the conditioning box (red). As more points are generated, the predicted boxes become more stable and better aligned with the ground truth. The CFSC metric captures this consistency, with higher mean IoU indicating greater geometric fidelity under controlled conditions.
    }
    \label{supp_fig:glenet_sample}
\end{figure}

\vspace{0.5mm}
\noindent\textbf{$\triangleright${ Conditional Foreground Spatial Consistency (CFSC).}}
\label{metric_cfsc}
CFSC evaluates the geometric fidelity of generated foreground point clouds under conditional generation. Inspired by GLENet~\cite{zhang2023glenet} and illustrated in Fig.~\ref{supp_fig:glenet_sample}, it measures how well the generated points reconstruct the conditioning 3D bounding box.

Specifically, for models conditioned on GT 3D boxes, we first extract foreground points from the generated scene within the conditioning box. A conditional variational autoencoder (CVAE), pretrained on real data, is then used to regress 3D box parameters from these cropped point clouds. The CVAE is trained to align the latent distribution of foreground points alone with the joint distribution conditioned on both points and class/box labels, following the GLENet formulation.

The CFSC score is computed as the average IoU between multiple sampled boxes and the ground-truth box:
\begin{equation}
    \mathrm{CFSC} = \frac{1}{N} \sum_{j=1}^{N} \left( \frac{1}{S} \sum_{s=1}^{S} \mathrm{IoU}\big(\hat{b}{j}^{(s)}, b{j}^{\mathrm{GT}}\big) \right)\,,
\label{eq:cfsc}
\end{equation}
where $N$ is the number of evaluated objects, $S$ is the number of samples per object, $\hat{b}_{j}^{(s)}$ is the $s$-th predicted box from the CVAE, and $b_{j}^{\mathrm{GT}}$ is the corresponding ground-truth box.

A higher CFSC score indicates stronger geometric consistency between generated points and the conditioning box, reflecting the model’s controllability and fidelity under conditional settings.

\subsubsection{Temporal-Level Evaluations.}
While the preceding metrics evaluate the quality of individual frames and foreground objects, our temporal-level metrics are designed to assess the \textbf{geometric and motion consistency} of generated \textbf{4D LiDAR sequences}.

\vspace{0.5mm}
\noindent\textbf{$\triangleright${ Temporal Transformation Consistency Error (TTCE).}}
\label{metric_ttce}
To assess temporal consistency in generated sequences, we use a point cloud registration algorithm (\emph{i.e.}, Iterative Closest Point (ICP)) to estimate the rigid transformation (rotation and translation) between consecutive frames. These estimated transformations are then compared against ground-truth transformations, which are available for each sequence.

Let $\mathbf{T}_t^{\mathrm{gt}} = [\mathbf{R}_t^{\mathrm{gt}} | \mathbf{t}_t^{\mathrm{gt}}]$ be the ground-truth transformation from frame $t$ to $t+1$, and $\mathbf{T}_t^{\mathrm{pred}} = [\mathbf{R}_t^{\mathrm{pred}} | \mathbf{t}_t^{\mathrm{pred}}]$ the transformation estimated from the generated point clouds via ICP. The TTCE is computed as:
\begin{equation}
    \mathrm{TTCE}{\mathrm{rot}} = \frac{1}{T-1} \sum{t=1}^{T-1} \left| \mathbf{R}_t^{\mathrm{pred}} (\mathbf{R}_t^{\mathrm{gt}})^\top - \mathbf{I} \right|_F\,,
\end{equation}
\begin{equation}
    \mathrm{TTCE}{\mathrm{trans}} = \frac{1}{T-1} \sum{t=1}^{T-1} \left| \mathbf{t}_t^{\mathrm{pred}} - \mathbf{t}_t^{\mathrm{gt}} \right|_2\,,
\label{eq:ttce}
\end{equation}
where $\|\cdot\|_F$ denotes the Frobenius norm, and $T$ is the total number of frames in the sequence.
Lower TTCE values indicate better alignment with ground truth, reflecting higher temporal coherence and more realistic motion in the generated sequence.

\vspace{0.5mm}
\noindent\textbf{$\triangleright${ Chamfer Temporal Consistency (CTC).}}
\label{metric_ctc}
CTC is a geometric-level metric for evaluating temporal consistency in generated sequences. It measures how smoothly point clouds evolve over time by computing the Chamfer Distance (CD) between consecutive frames after aligning them using ground-truth transformations.

Let $\mathcal{P}_t$ and $\mathcal{P}_{t+1}$ be the generated point clouds at frames $t$ and $t+1$. We align $\mathcal{P}_{t+1}$ to the coordinate frame of $t$ using the ground-truth transformation $\mathbf{T}_t^{\mathrm{gt}} = [\mathbf{R}_t^{\mathrm{gt}} | \mathbf{t}_t^{\mathrm{gt}}]$:
\begin{equation}
    \tilde{\mathcal{P}}_{t+1} = (\mathbf{R}_t^{\mathrm{gt}})^{-1} (\mathcal{P}_{t+1} - \mathbf{t}_t^{\mathrm{gt}})\,.
\end{equation}
The Chamfer Distance between $\mathcal{P}_t$ and the aligned $\tilde{\mathcal{P}}_{t+1}$ is then computed as:
\begin{equation}
\begin{split}
    \mathrm{CD}(\mathcal{P}_t, \tilde{\mathcal{P}}_{t+1}) = &\frac{1}{|\mathcal{P}_t|} \sum_{x \in \mathcal{P}_t} \min_{y \in \tilde{\mathcal{P}}_{t+1}} \|x - y\|_2^2 + \\
    &\frac{1}{|\tilde{\mathcal{P}}_{t+1}|} \sum_{y \in \tilde{\mathcal{P}}_{t+1}} \min_{x \in \mathcal{P}_t} \|y - x\|_2^2\,.
\end{split}
\end{equation}
The final CTC score is reported as the mean Chamfer Distance across all consecutive frame pairs:
\begin{equation}
    \mathrm{CTC} = \frac{1}{T-1} \sum_{t=1}^{T-1} \mathrm{CD}(\mathcal{P}_t, \tilde{\mathcal{P}}_{t+1})\,.
\label{eq:ctc}
\end{equation}
Lower CTC values indicate smoother transitions and stronger temporal coherence in the 4D LiDAR sequences.

\subsubsection{Layout-Level Evaluations.}
Layout-level evaluation focuses on the \textbf{spatial arrangement} and \textbf{motion realism} of objects at the scene graph level. Since each scene graph node corresponds to a generated 3D bounding box and its associated trajectory, these metrics comprehensively assess \textbf{inter-object spatial relationships} and \textbf{dynamic interactions}.

In our experiments, we convert dataset annotations into ground-truth scene graphs and generate 10,000 new layouts for evaluation. All layout-level metrics are computed on these generated samples, using the corresponding ground-truth scene graphs as reference for each scenario.

\vspace{0.5mm}
\noindent\textbf{$\triangleright${ Spatial Consistency Rate (SCR)}}
\label{metric_scr}
measures the proportion of object pairs whose spatial relationships in the generated layout match those specified in the ground-truth scene graph. Let $N_{\mathrm{pairs}}$ be the total number of object pairs, and $N_{\mathrm{consistent}}$ the number of pairs with consistent spatial relations:
\begin{equation}
    \mathrm{SCR} = \frac{N_{\mathrm{consistent}}}{N_{\mathrm{pairs}}}\,.
\label{eq:scr}
\end{equation}

\vspace{0.5mm}
\noindent\textbf{$\triangleright${ Motion State Consistency Rate (MSCR)}}
\label{metric_mscr}
quantifies the proportion of generated object trajectories whose motion state (\emph{e.g.}, moving, stopped, turning) matches the state specified in the scene graph. Let $N_{\mathrm{traj}}$ be the number of object trajectories and $N_{\mathrm{match}}$ those matching the ground-truth motion state:
\begin{equation}
    \mathrm{MSCR} = \frac{N_{\mathrm{match}}}{N_{\mathrm{traj}}}\,.
\label{eq:mscr}
\end{equation}

\vspace{0.5mm}
\noindent\textbf{$\triangleright${ Box Collision Rate (BCR)}}
\label{metric_bcr}
BCR computes the proportion of object box pairs that collide (\emph{i.e.}, their 3D IoU $>0$) across all frames and all box pairs. Let $\mathcal{F}$ be the set of all frames, and $\mathcal{P}_t$ the set of object pairs in frame $t$. Then:
\begin{equation}
    \mathrm{BCR} = \frac{\sum_{t\in\mathcal{F}} \sum_{(i,j)\in\mathcal{P}_t} \mathbb{I}(\mathrm{IoU}(b_i^{(t)}, b_j^{(t)}) > 0)}{\sum_{t\in\mathcal{F}} |\mathcal{P}_t|}\,.
\label{eq:bcr}
\end{equation}

\vspace{0.5mm}
\noindent\textbf{$\triangleright${ Trajectory Collision Rate (TCR)}}
\label{metric_tcr}
TCR calculates the proportion of all trajectory pairs whose associated boxes overlap (\emph{i.e.}, collide) at any time step in the sequence. Let $\mathcal{Q}$ be the set of all unordered trajectory pairs $(m,n)$, and $\mathbb{I}_{\mathrm{traj\_collide}}(m,n)$ be 1 if their boxes collide at any time $t$, else 0. Then:
\begin{equation}
    \mathrm{TCR} = \frac{\sum_{(m,n)\in\mathcal{Q}} \mathbb{I}\left(\exists t: \mathrm{IoU}(b_m^{(t)}, b_n^{(t)}) > 0 \right)}{|\mathcal{Q}|}\,.
\label{eq:tcr}
\end{equation}

\subsection{Summary of Generations Baselines}
\label{baseline}
Table~\ref{supp_tab:baselines} summarizes all baseline models used for comparison in this work. The selected methods span both unconditional and conditional generative paradigms, encompassing a variety of input representations, model architectures, and generative strategies. For each baseline, we report its publication venue, core generation mechanism, and average per-frame inference time.

\vspace{0.5mm}
\noindent\textbf{$\triangleright${ Training Free Sequence Generation with LiDARCrafter.}} 
To support our hypothesis that point cloud data is inherently well-suited for autoregressive generation in 4D, we introduce a training-free baseline that incorporates historical scene information when generating future frames -- without retraining the model.

Starting from an initial generated frame, we apply a pretrained point cloud segmentation model to remove ground points, isolating dynamic and foreground elements. For each subsequent frame, we transform the previously generated scene into the current frame’s coordinate system using our scene editing pipeline, applying a spatial mask to retain only relevant regions. The generative model then synthesizes the new frame conditioned on this transformed historical context, ensuring both temporal coherence and physical plausibility across the sequence.

This baseline allows us to explicitly exploit autoregressive dependencies in 4D point cloud generation and compare its performance against unconditional or single-frame generation methods.

%% file: supp_tables/baselines.tex
    \centering
    \caption{
    Summary of baseline LiDAR-based generative models used for comparisons in this work. For each model, we report the method name, publication venue, data representation, generative paradigm, model architecture, and average inference time per frame (I.T., in seconds). DM denotes the diffusion model.
    }
    \vspace{-0.3cm}
    \resizebox{\linewidth}{!}{
    \begin{tabular}{c|r|r|c|c|c|c}
    \toprule
    \textbf{Base}& \textbf{Method}       & \textbf{Venue} 
      & \textbf{Representation}
      & \textbf{Generator}
      & \textbf{Architecture}
      & \textbf{I.T. (s)} \\
    \midrule\midrule
    \multirow{3.1}{*}{\rotatebox{90}{\small \textbf{Uncond.}}}
    & LiDARGen \cite{zyrianov2022learning}     & ECCV'22 
      & Range Image  & Energy-Based 
      & U-Net  & $67$ \\
    &LiDM \cite{ran2024towards}         & CVPR'24 
      & Range Image  & Latent DM 
      & Curve U-Net  & $3.5$ \\
    &R2DM \cite{nakashima2024lidar}        & ICRA'24 
      & Range Image  & DM 
      & Curve U-Net  & $3.5$ \\
    \midrule
    \multirow{4.5}{*}{\rotatebox{90}{\textbf{Cond.}}}
    &UniScene \cite{li2025uniscene}      & CVPR'25 
      & Voxel  & Volume Rendering 
      & Sparse U-Net  & $2.1$ \\
    &OpenDWM \cite{opendwm}      & CVPR'25 
      & Voxel  & Latent DM 
      & U-Net  & $12.1$  \\
    &OpenDWM-DiT \cite{opendwm}  & CVPR'25 
      & Voxel  & Latent DM 
      & Transformer  & $11.4$  \\
    \cmidrule{2-7}
    &\textbf{LiDARCrafter} & \textbf{Ours} 
      & Range Image  & DM
      & Curve U-Net  & $3.6$ \\
    \bottomrule
    \end{tabular}}
    \label{supp_tab:baselines}

%% file: supp_tables/layout_gen.tex
    \centering
    \caption{
    Comparison of joint \textit{vs.} split generation architectures for 4D layout generation. We report spatial consistency rate (SCR), motion state consistency rate (MSCR), box collision rate (BCR), trajectory collision rate (TCR), and average inference time per frame (I.T., in ms).
    }
    \vspace{-0.2cm}
    \resizebox{\linewidth}{!}{
    \begin{tabular}{c|c|ccccc}
    \toprule
    \# & \textbf{Config} 
      & \textbf{SCR}$\uparrow$ 
      & \textbf{MSCR}$\uparrow$ 
      & \textbf{BCR}$\downarrow$ 
      & \textbf{TCR}$\downarrow$ 
      & \textbf{I.T. (ms)}$\downarrow$ \\
    \midrule\midrule
    $$1$$ & Joint  
      & $\mathbf{91.77}$
      & $$87.41$$ 
      & $$10.57$$ 
      & $$35.43$$ 
      & $\mathbf{150}$ \\
    $$2$$ & Split  
      & $$90.72$$ 
      & $\mathbf{89.24}$ 
      & $\mathbf{9.79}$ 
      & $\mathbf{17.29}$
      & $$270$$ \\
    \bottomrule
    \end{tabular}}
    \label{supp_tab:config_joint_split}

%% file: supp_tables/notations.tex
\begin{table}[t]
\centering
\caption{Summary of notations defined in this work.}
\vspace{-0.2cm}
\resizebox{.8\linewidth}{!}{
\begin{tabular}{c|l}
    \toprule
    \textbf{Notation} & \textbf{Definition}
    \\\midrule\midrule
    $\mathcal{P}$ & LiDAR point cloud sequence
    \\
    $\mathbf{P}$ & LiDAR point cloud frame
    \\
    {$t$} & Timestamp of a LiDAR sequence
    \\
    {$N$} & Number of points 
    \\
    {$\Pi$} & Projection function to range view
    \\
    {$H,W$} & Shape of range image
    \\
    {$\mathcal{G}$} & Scene graph
    \\
    {$\mathcal{V}$} & Nodes set of scene graph
    \\
    {$\mathcal{E}$} & Edges set of scene graph
    \\
    {$v_i$} & A node of scene graph
    \\
    {$e_{i\rightarrow j}$} & An edge of scene graph
    \\
    {$M$} & Number of nodes in a scene graph
    \\
    {$\tau$} & Timestamp in a diffusion process
    \\
    {$\mathbf{d}_{\tau}$} & Noisy sample in a diffusion process
    \\
    {$\varepsilon_\theta$} & Denoiser in a diffusion model
    \\
    {$c$} & Conditions in a diffusion process
    \\
    {$\mathbf{b}_i$} & Generated 3D bounding box of a node
    \\
    {$(x_i, y_i, z_i)$} & Center position of 3D bounding box
    \\
    {$(w_i, l_i, h_i)$} & Width, length and height of box 
    \\
    {$\psi_i$} & Yaw angle of box
    \\
    {$\boldsymbol{\delta}_i$} & Generated trajectory of a node
    \\
    {$\Delta x_i^t$} & horizontal planar displacement of object
    \\
    {$\Delta y_i^t$} & vertical planar displacement of object
    \\
    {$\mathbf{p}_i$} & Generated object point cloud of a node
    \\
    {$c_i$} & Category of a node
    \\
    {$s_i$} & Motion state of a node
    \\
    {$\mathbf{h}^{(\ell)}_{v_i},\mathbf{h}^{(\ell)}_{e_{i\rightarrow j}}$} & Node and edge feature
    \\
    {$\mathcal{L}$} & Loss function
    \\
    {$\mathcal{O}_i$} & Learned layout of a node
    \\
    {$\Phi$} &  MLP layers
    \\
    {$L$} & Number of layers in a TripletGCN model
    \\
    {$o$} & Task for each diffusion branch
    \\
    {$\mathbf{I}$} & Task for each diffusion branch
    \\ 
    {$\mathbf{m}$} & Binary mask projected by edited bounding boxes
    \\ 
    {$\mathbf{B}$} & Background points back-projected from a range image
    \\ 
    {$\mathbf{F}$} & Foreground points back-projected from a range image
    \\ 
    {$\mathbf{u}$} & Translation from the world origin
    \\ 
    {$\mathbf{R}$} & Rotation matrix
    \\ 
    {$\mathbf{G}$} & Homogeneous transformation matrix
    \\ 

    \bottomrule
\end{tabular}}
\label{supp_tab: notations}
\end{table}

%% file: supp_tables/metrics.tex
\begin{table*}[!ht]
    \centering
    \vspace{-0.3cm}
    \caption{Summary of evaluation metrics used for LiDAR scene, object, sequence, and layout generation quality.}
    \renewcommand\arraystretch{1.3}
    \resizebox{\linewidth}{!}{
    \begin{tabular}{|>{\centering\arraybackslash}m{0.8cm}|c|>{\centering\arraybackslash}m{3.9cm}|m{8.2cm}|c|}
        \hline
        \textbf{Level} & \textbf{Abbr.} & \textbf{Full Name} & \textbf{Motivation} & \textbf{Reference} \\
        \hline\hline
        \multirow{6}{*}{\rotatebox{90}{Static Scene}}
            & \textbf{FRD} & Fréchet Range Distance & Measures the feature distributional similarity between generated and real range images. Lower is better. & Eq.~(\ref{eq:frd})
            \\
        \cline{2-5}
            & \textbf{FPD} & Fréchet Point Distance & Evaluates the feature similarity between generated and real 3D point cloud distributions. Lower is better.& Eq.~(\ref{eq:fpd}) 
            \\
        \cline{2-5}
            & \textbf{JSD} & \small{Jensen–Shannon Divergence} & Quantifies the difference between point cloud distributions in a probabilistic manner. Lower is better. & Eq.~(\ref{eq:jsd}) 
            \\
        \cline{2-5}
            & \textbf{MMD} & \small{Maximum Mean Discrepancy} & Measures the distance between generated and real distributions to evaluate diversity and coverage. Lower is better.& Eq.~(\ref{eq:mmd})
            \\
        \hline

        \multirow{9}{*}{\rotatebox{90}{Foreground Object} }
            & \textbf{FDC} & \small{Foreground Detection Confidence} & Assesses the confidence of detected objects from generated samples with a pretrained detector. Higher is better.& Eq.~(\ref{eq:fdc})
            \\
        \cline{2-5}
            & \textbf{CDA} & \small{Conditioned Detection Accuracy} & Applies a pretrained 3D detector to generated point clouds (under box-conditioned generation) and computes standard detection metrics, providing a comprehensive assessment of geometric and semantic fidelity. Higher is better.& -
            \\
        \cline{2-5}
            & \textbf{CFCA} & \small{Conditioned Foreground Classification Accuracy} & Evaluates semantic consistency by classifying generated objects using a pretrained classifier. Higher is better.& Eq.~(\ref{eq:cfca})
            \\
        \cline{2-5}
            & \textbf{CFSC} & \small{Conditional Foreground Spatial Consistency} & A conditional variational autoencoder regresses 3D box parameters from generated foreground points; the mean IoU between predicted and GT boxes quantifies geometric consistency under conditioning. Higher is better.& Eq.~(\ref{eq:cfsc}) 
            \\
        \hline

        \multirow{4}{*}{\rotatebox{90}{\makecell{Temporal \\ Sequence}}}
            & \textbf{TTCE} & \small{Temporal Transformation Consistency Error} & Measures temporal consistency by registering generated point clouds between frames and comparing the estimated transformation matrix to ground truth. Lower error indicates better temporal alignment. Lower is better.& Eq.~(\ref{eq:ttce})
            \\
        \cline{2-5}
            & \textbf{CTC} & \small{Chamfer Temporal Consistency} & Evaluates temporal consistency by aligning backgrounds of consecutive generated point clouds and computing Chamfer Distance. Lower is better. & Eq.~(\ref{eq:ctc})
            \\
        \hline

        \multirow{7}{*}{\rotatebox{90}{Layout Generation}}
            & \textbf{SCR} & \small{Spatial Consistency Rate} & Measures the proportion of generated 3D boxes whose spatial relationships match those specified in the scene graph. Higher is better.& Eq.~(\ref{eq:scr}) 
            \\
        \cline{2-5}
            & \textbf{MSCR} & \small{Motion State Consistency Rate} & Measures the proportion of generated object trajectories whose motion states are consistent with the scene graph specifications. Higher is better.& Eq.~(\ref{eq:mscr}) 
            \\
        \cline{2-5}
            & \textbf{BCR} & \small{Box Collision Rate} & Calculates the proportion of object pairs whose generated 3D bounding boxes overlap in the scene. Lower is better. & Eq.~(\ref{eq:bcr}) 
            \\
        \cline{2-5}
            & \textbf{TCR} & \small{Trajectory Collision Rate} & Calculates the proportion of object trajectory pairs whose predicted 3D trajectories collide. Lower is better. & Eq.~(\ref{eq:tcr})
            \\
        \hline
    \end{tabular}
    }
    \label{tab:metric_summary}
\end{table*}

%% file: supp_sections/2_addtional_exp.tex
\section{Additional Experimental Analyses}
\label{sec_supp: addtional_exp}
In this section, we present additional quantitative and qualitative results to further demonstrate the effectiveness and advantages of \textbf{LiDARCrafter}.

\subsection{Additional Quantitative Results}
\noindent\textbf{Ablation on Effectiveness of Tri-Branch Architecture for Layout Generating.}
To evaluate the effectiveness of our tri-branch architecture for 4D layout generation, we conduct an ablation study comparing our default split generation strategy (separate box and trajectory branches) with a joint generation approach that predicts boxes and trajectories simultaneously.

As shown in Table~\ref{supp_tab:config_joint_split}, the split architecture achieves superior motion and collision consistency (higher MSCR, lower BCR and TCR), while the joint architecture offers a slight improvement in spatial consistency (SCR) and inference speed. These results validate the benefits of decoupling static and dynamic layout generation in complex scene modeling.

\begin{figure*}[t]
    \centering
    \includegraphics[width=\textwidth]{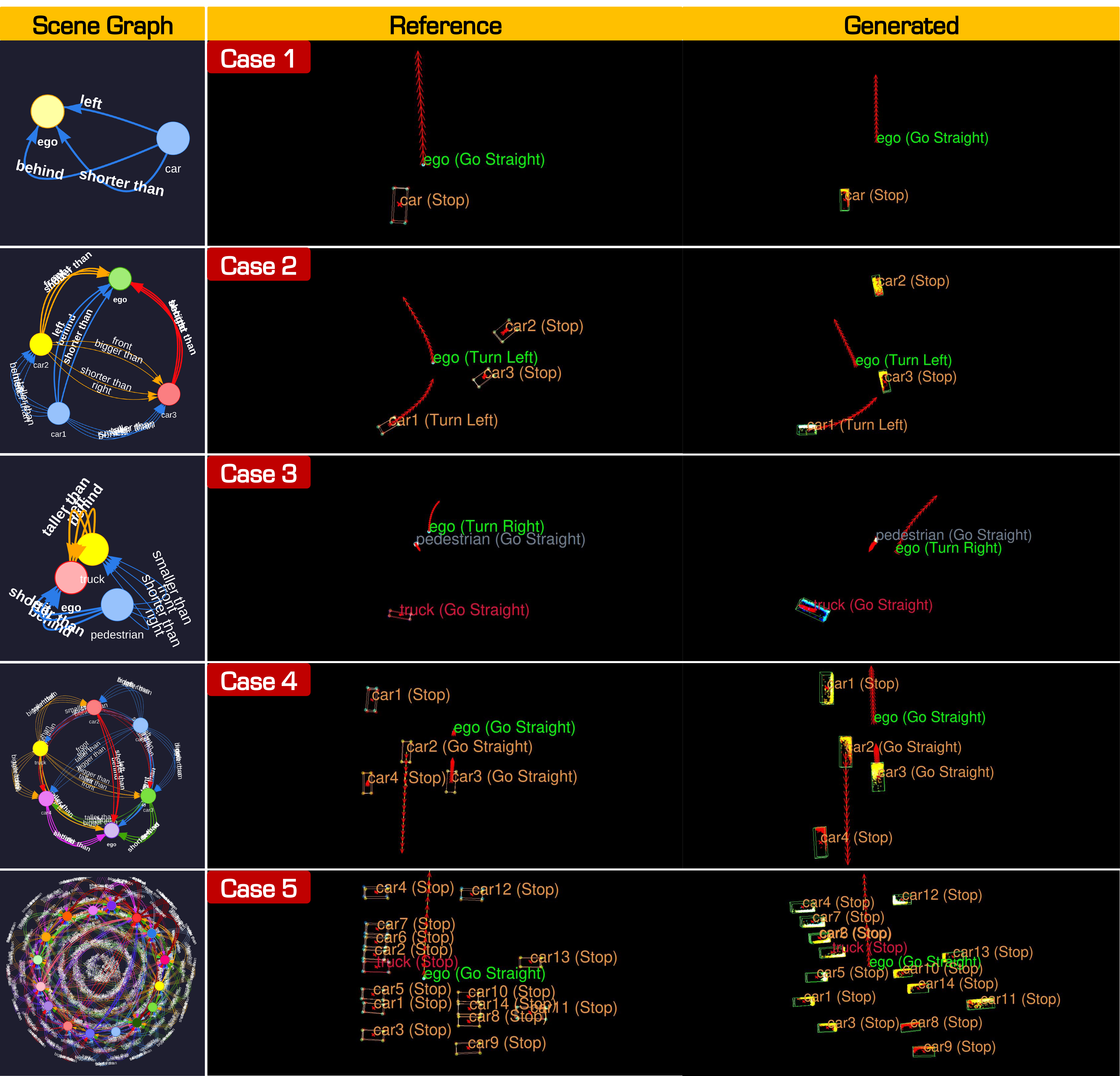}
    \vspace{-0.5cm}
    \caption{
        Visualization of four 4D layout generation cases. For each case, we show (from left to right): the input scene graph, the ground-truth (GT) object layout, and the layout generated by LiDARCrafter. Our method produces layouts that accurately match the required spatial relationships and motion states defined by the scene graph, while successfully avoiding collisions between boxes and trajectories. Even in cases with numerous objects and complex spatial constraints (case 5), the generated results maintain high spatial and temporal consistency.
    }
    \label{supp_fig:4d_layout}
\end{figure*}

\subsection{Additional Qualitative Results}
\noindent\textbf{Spatially and Temporally Consistent 4D Layout Generation.}
Figure~\ref{supp_fig:4d_layout} visualizes four representative 4D layout generation cases. For each case, we present the input scene graph, the ground-truth (GT) layout, and the corresponding layout generated by LiDARCrafter. The visualizations demonstrate that our approach produces object arrangements and motion patterns that are highly consistent with the spatial relationships and motion states specified by the scene graph. Notably, the generated results avoid box and trajectory collisions even in scenarios with many objects and complex spatial relations (see case 5), highlighting the effectiveness of our tri-branch layout generation and physical constraint mechanisms.

\begin{figure*}[t]
    \centering
    \includegraphics[width=.85\textwidth]{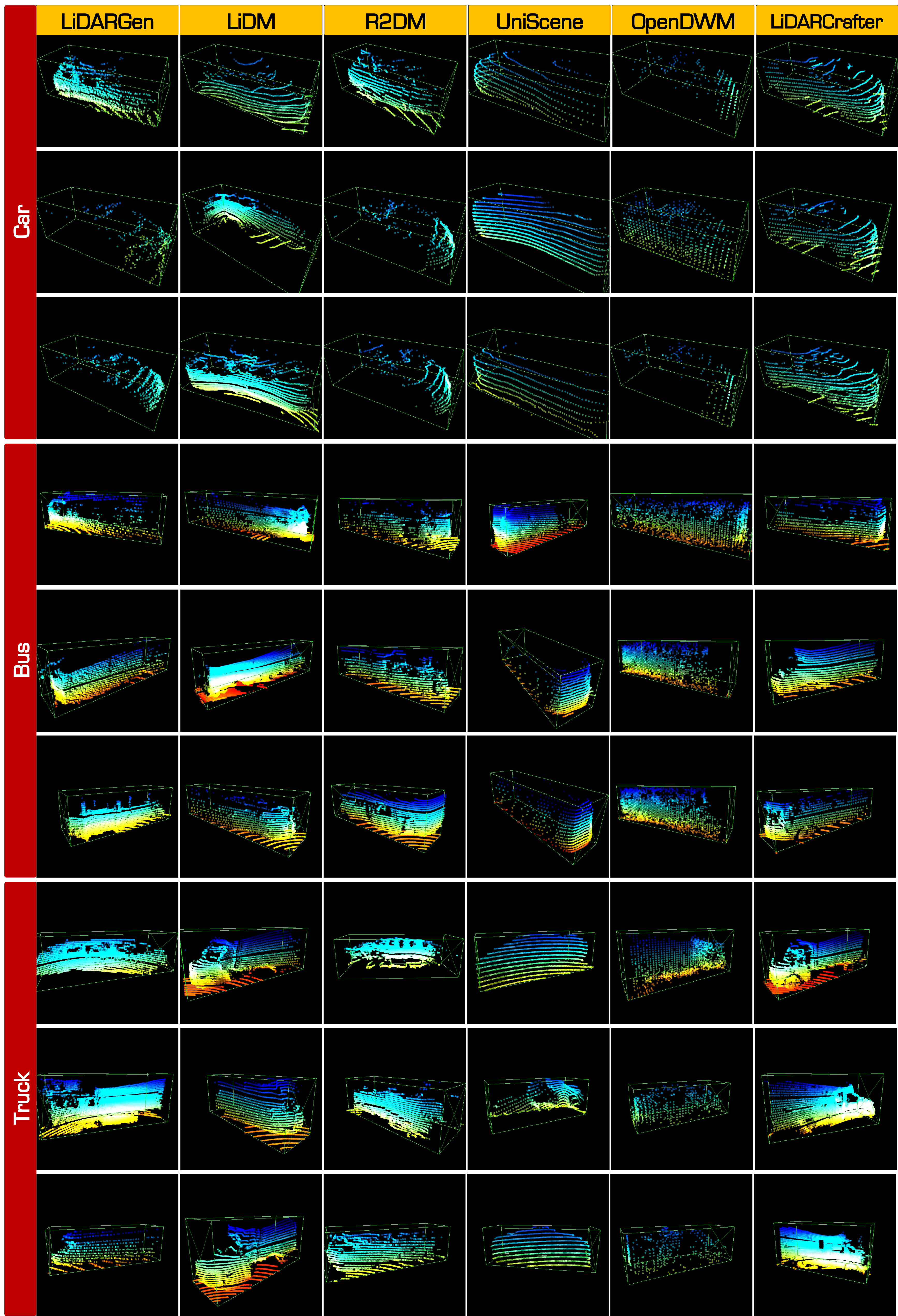}
    \vspace{-0.1cm}
    \caption{
    Qualitative comparison of generated foreground objects across different methods. For each category (car, bus, truck), three high-confidence examples are shown, extracted using a pretrained 3D detector. LiDARCrafter produces the cleanest and most geometrically faithful foreground objects, while alternative methods either exhibit noisy outputs (LiDARGen), suffer from blurry local structures (LiDM, R2DM), or fail to capture LiDAR scan patterns (UniScene, OpenDWM).
    }
    \label{supp_fig:fg_objects}
\end{figure*}

\noindent\textbf{Comparison of Foreground Object Quality Across Methods.}
To evaluate the geometric quality of generated foreground objects, we use a pretrained 3D object detector to extract high-confidence object instances from the generated scenes of each method. We focus on three representative categories: car, bus, and truck, visualizing three examples per class for each method. As shown in Figure~\ref{supp_fig:fg_objects}, LiDARCrafter generates foreground objects with the cleanest shapes and the most accurate geometric structures, closely resembling real LiDAR scans. In contrast, LiDARGen produces objects with significant noise despite preserving basic outlines. LiDM and R2DM yield moderately improved results, but local structures are often blurry or indistinct. Notably, UniScene and OpenDWM fail to capture the characteristic LiDAR scanning patterns, resulting in unrealistic foreground point clouds.

\begin{figure*}[t]
    \centering
    \includegraphics[width=.92\textwidth]{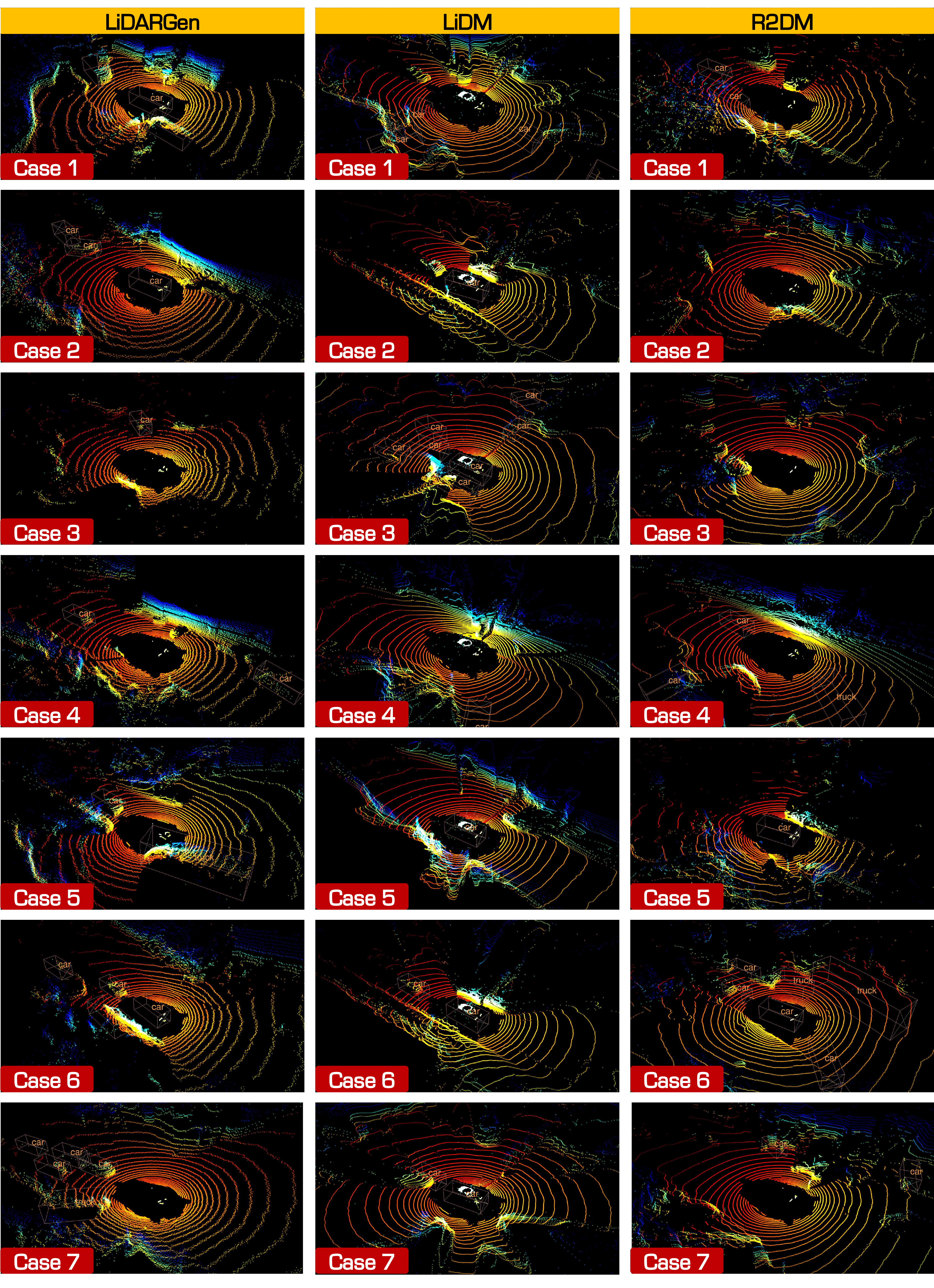}
    \vspace{-0.25cm}
    \caption{
        Additional generation examples for unconditional methods. Detected objects are highlighted using a pretrained VoxelRCNN ($>0.5$ confidence). LiDARCrafter maintains realistic LiDAR patterns and accurate object structures across frames.
    }
    \label{fig:single_frame_uncond}
\end{figure*}

\begin{figure*}[t]
    \centering
    \includegraphics[width=.92\textwidth]{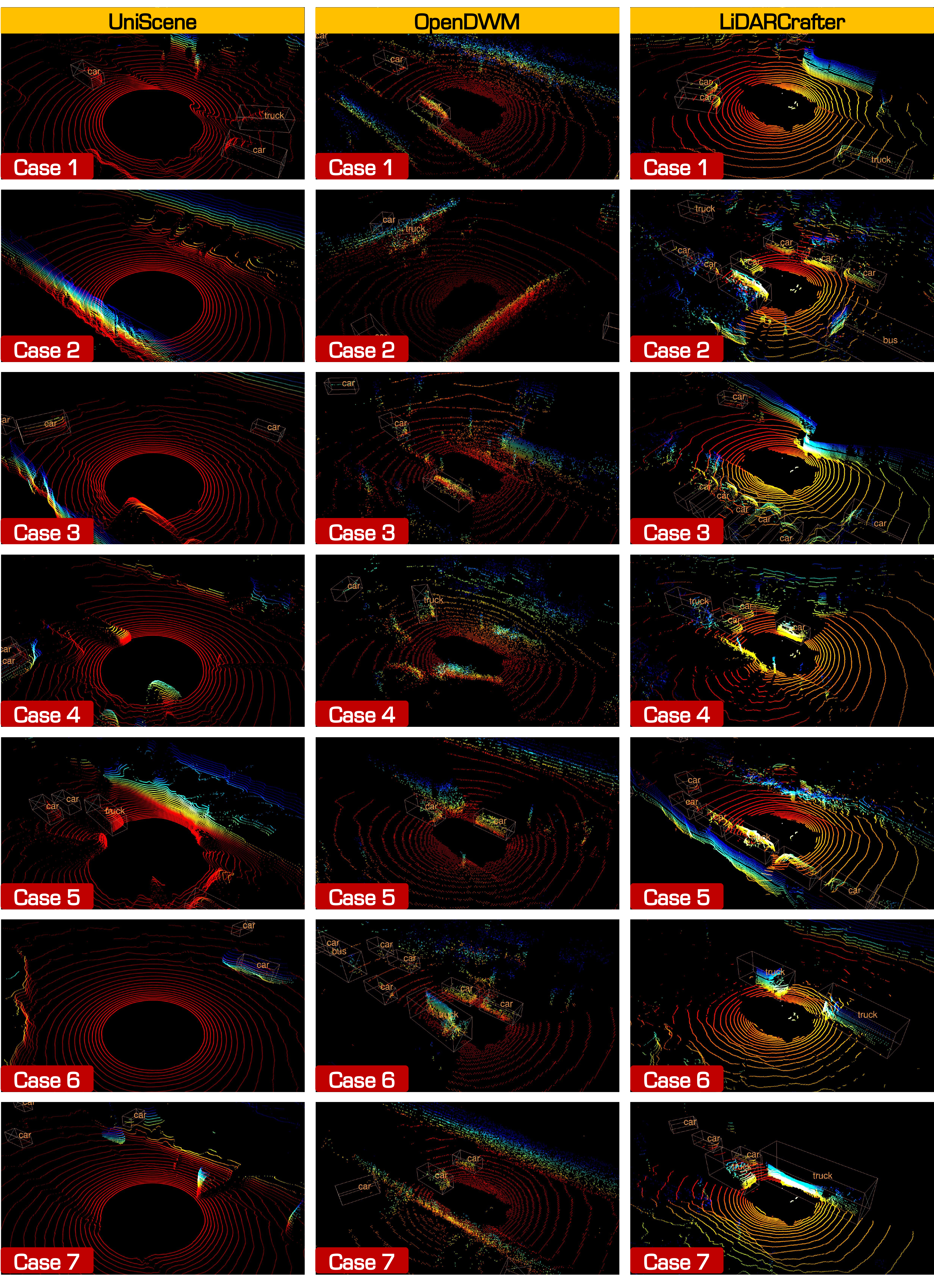}
    \vspace{-0.25cm}
    \caption{
        Additional generation examples for conditional methods, with VoxelRCNN detections. Our method achieves both realistic scanning patterns and robust preservation of foreground geometry, enabling strong downstream performance.
    }
    \label{fig:single_frame_cond}
\end{figure*}

\noindent\textbf{Frame Generation Results Across Methods.}
We further present qualitative results comparing sequence generation methods under both unconditional and conditional settings. Figure~\ref{fig:single_frame_uncond} shows examples from unconditional generation methods, with detected objects visualized using a pretrained VoxelRCNN (confidence $>0.5$). Figure~\ref{fig:single_frame_cond} displays results from conditional generation methods, likewise annotated with detector predictions. Across both settings, LiDARCrafter generates point clouds with patterns closely resembling real LiDAR scans, and uniquely preserves the geometric integrity of foreground objects throughout the sequence. This fidelity substantially improves the transferability of generated data to downstream tasks such as detection and tracking.

\begin{figure*}[t]
    \centering
    \includegraphics[width=.73\textwidth]{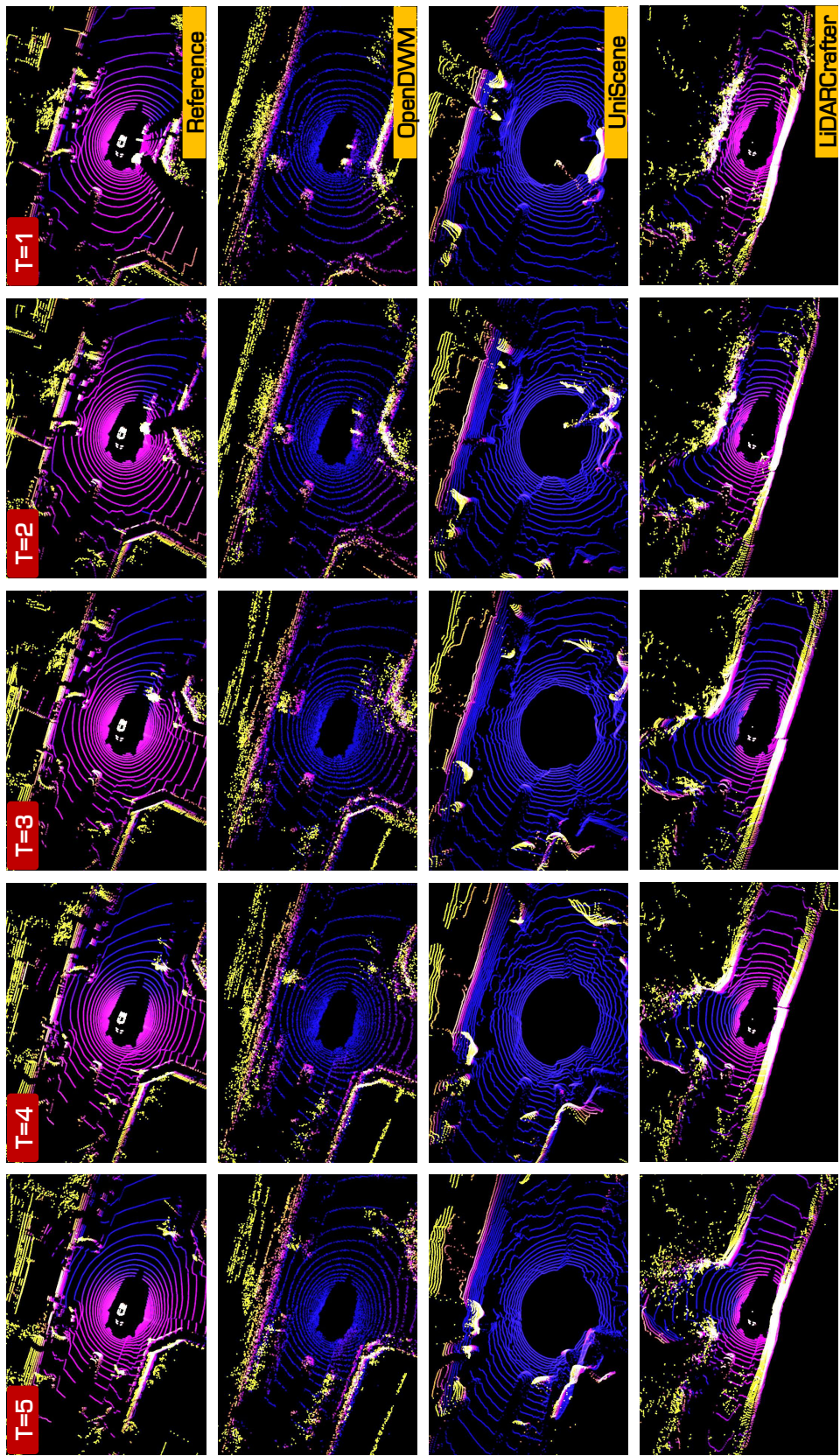}
    \vspace{-0.2cm}
    \caption{
    Qualitative comparison of $5$-frame ($2.5$s) sequence generation results. LiDARCrafter achieves both high temporal consistency and scanning patterns closest to ground truth. While voxel-based approaches (UniScene, OpenDWM) excel in temporal coherence, they severely disrupt the LiDAR-specific scan distribution, leading to unrealistic visual artifacts.
    }
    \label{supp_fig:cond_seq1}
\end{figure*}

\begin{figure*}[t]
    \centering
    \includegraphics[width=.73\textwidth]{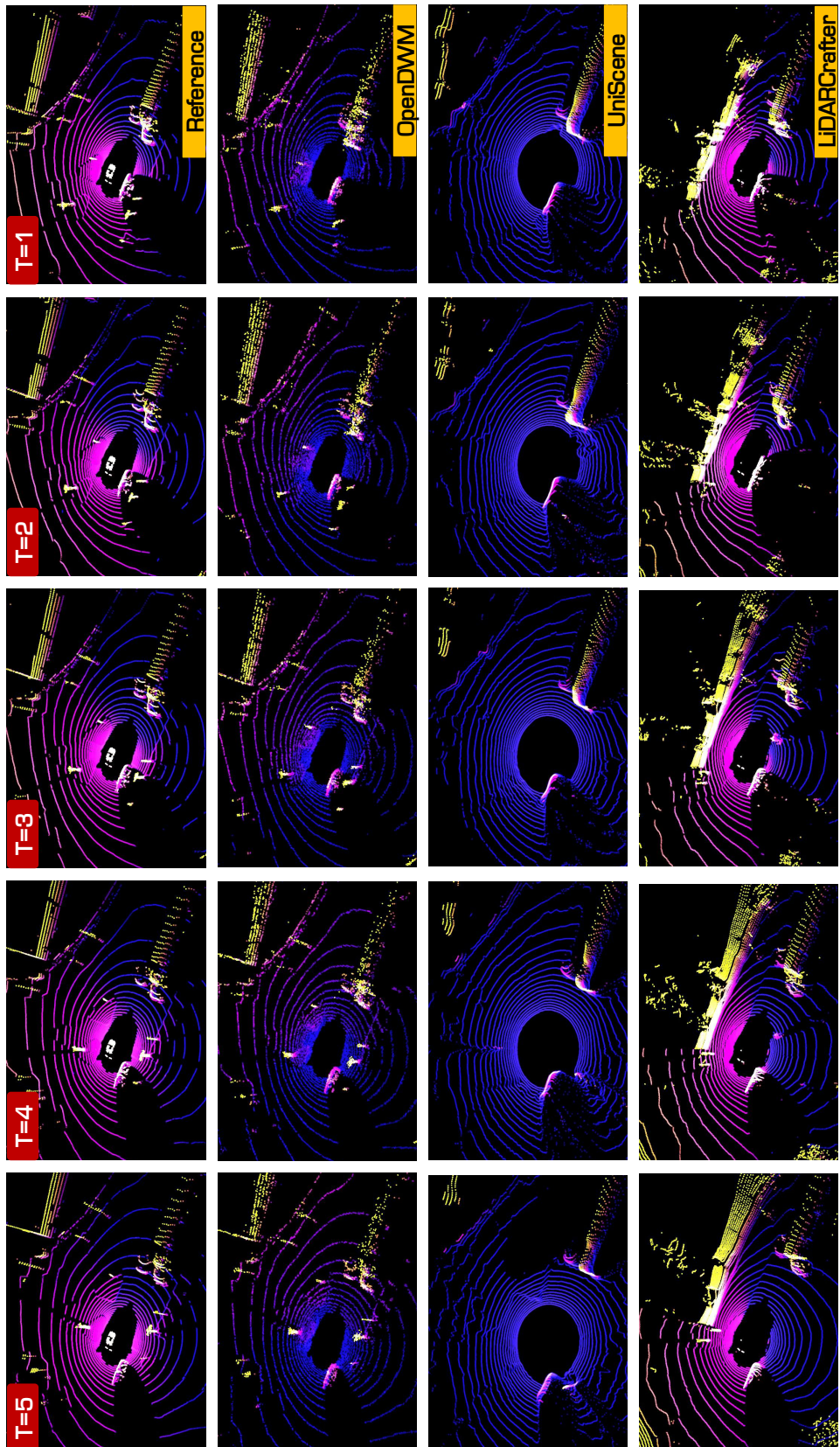}
    \vspace{-0.2cm}
    \caption{
    Qualitative comparison of $5$-frame ($2.5$s) sequence generation results. LiDARCrafter achieves both high temporal consistency and scanning patterns closest to ground truth. While voxel-based approaches (UniScene, OpenDWM) excel in temporal coherence, they severely disrupt the LiDAR-specific scan distribution, leading to unrealistic visual artifacts.
    }
    \label{supp_fig:cond_seq2}
\end{figure*}

\begin{figure*}[t]
    \centering
    \includegraphics[width=.73\textwidth]{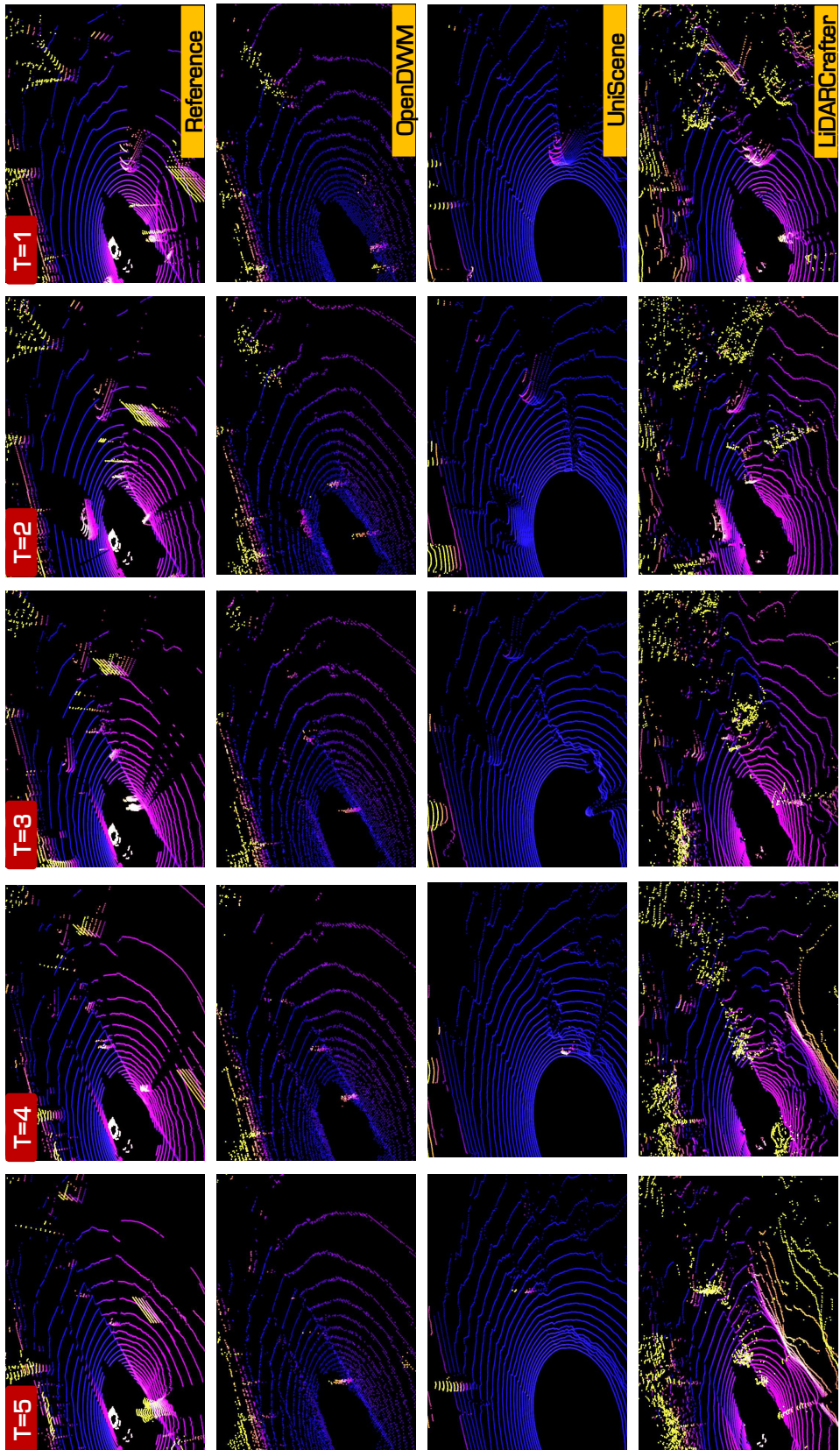}
    \vspace{-0.2cm}
    \caption{
    Qualitative comparison of $5$-frame ($2.5$s) sequence generation results. LiDARCrafter achieves both high temporal consistency and scanning patterns closest to ground truth. While voxel-based approaches (UniScene, OpenDWM) excel in temporal coherence, they severely disrupt the LiDAR-specific scan distribution, leading to unrealistic visual artifacts.
    }
    \label{supp_fig:cond_seq3}
\end{figure*}

\begin{figure*}[t]
    \centering
    \includegraphics[width=.73\textwidth]{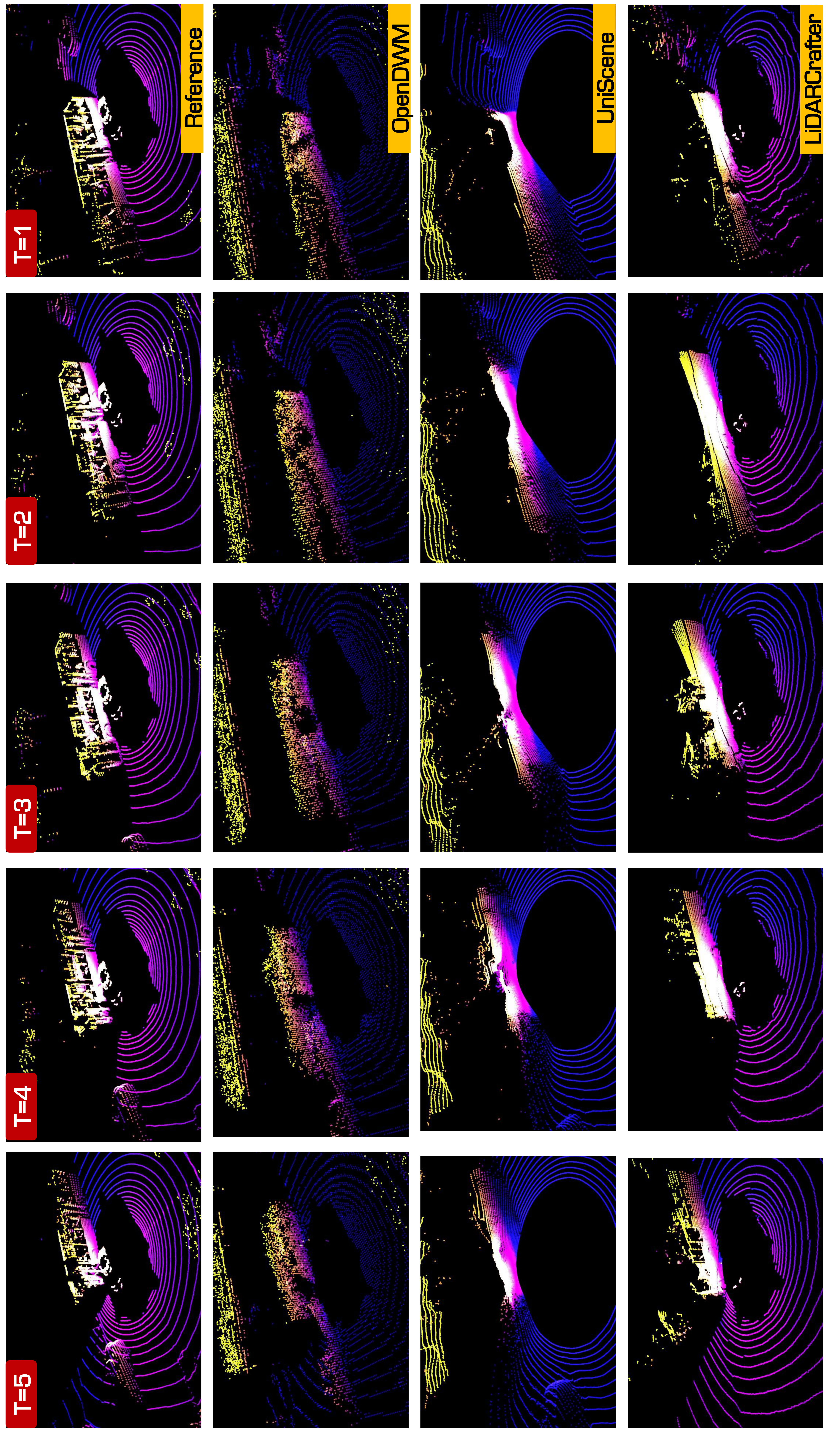}
    \vspace{-0.2cm}
    \caption{
    Qualitative comparison of $5$-frame ($2.5$s) sequence generation results. LiDARCrafter achieves both high temporal consistency and scanning patterns closest to ground truth. While voxel-based approaches (UniScene, OpenDWM) excel in temporal coherence, they severely disrupt the LiDAR-specific scan distribution, leading to unrealistic visual artifacts.
    }
    \label{supp_fig:cond_seq4}
\end{figure*}

\noindent\textbf{Additional Examples of Sequence Generation.}
To further highlight temporal consistency and realism, we present $5$-frame sequence generation results ($2.5$s interval, following the UniScene setting) for four different scenes. As visualized in Figures~\ref{supp_fig:cond_seq1}-\ref{supp_fig:cond_seq4}, LiDARCrafter not only maintains strong temporal consistency across all frames but also produces point distributions and scanning patterns that closely match ground-truth data. In comparison, voxel-based methods (such as UniScene and OpenDWM) preserve temporal alignment well due to the voxel representation, but fail to capture the characteristic LiDAR scan patterns -- resulting in unrealistic foreground details and scene textures.

\begin{figure*}[t]
    \centering
    \includegraphics[width=\textwidth]{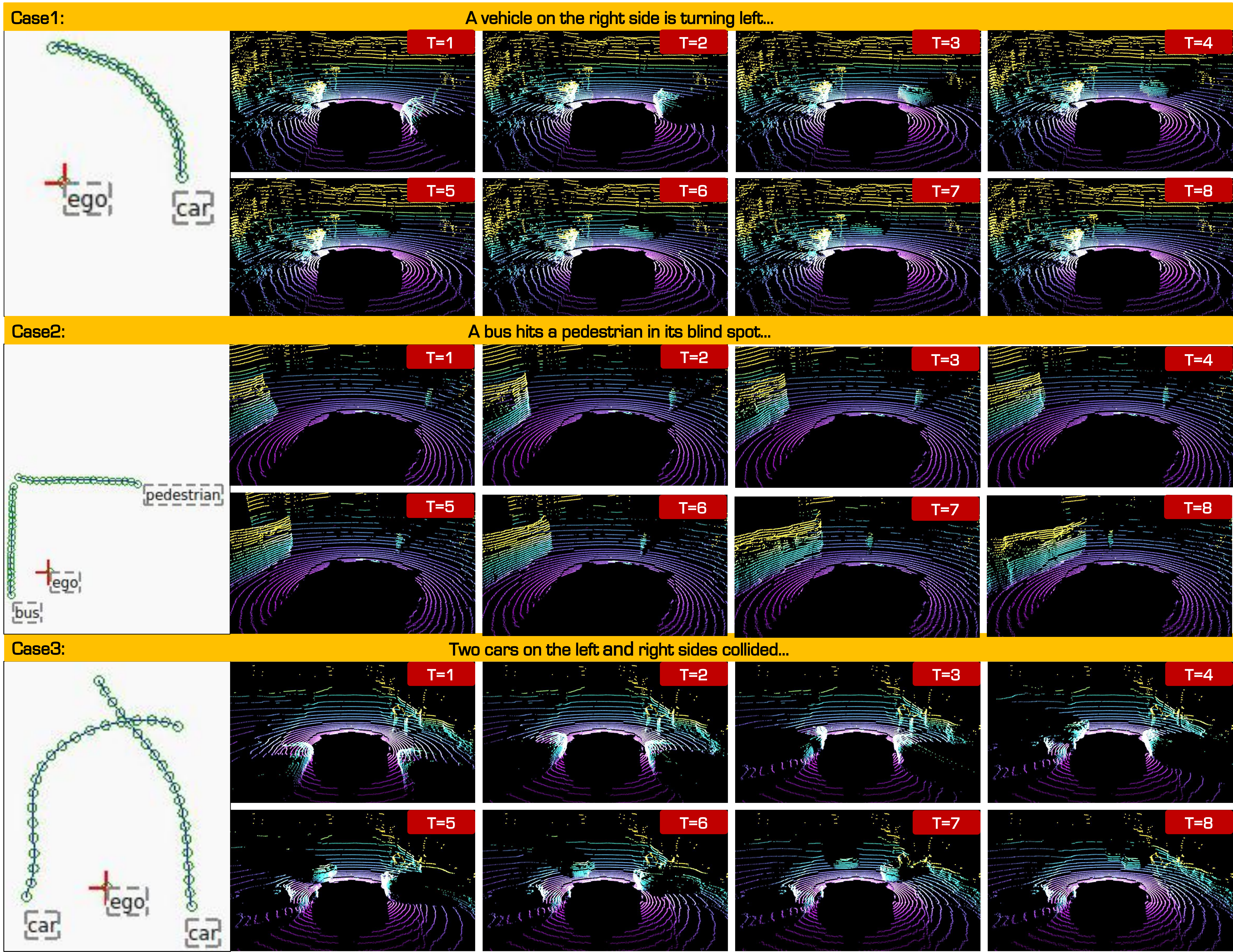}
    \vspace{-0.5cm}
    \caption{
        Visualization of three challenging corner-case driving scenarios generated by LiDARCrafter: (1) right-side overtaking followed by a sudden left turn; (2) a pedestrian in the bus’s blind spot; (3) two vehicles merging into the ego lane and colliding. In all cases, the generated point cloud sequences conform to the specified layouts, exhibit strong temporal consistency, and preserve fine-grained object geometry, demonstrating the model's creative flexibility for critical safety validation.
    }
    \label{supp_fig:corner_case}
\end{figure*}

\noindent\textbf{Corner-Case Scenario Generation Results.}
LiDARCrafter empowers users to design and generate any desired driving scenario, including complex and rare corner cases, by either manually specifying scene graphs via a user interface or parsing free-form natural language descriptions into structured layouts. In Figure~\ref{supp_fig:corner_case}, we showcase three challenging and safety-critical driving scenarios: (1) a vehicle overtakes on the right and rapidly turns left; (2) a pedestrian appears in the blind spot of a bus; and (3) two vehicles merge into the ego lane from both sides and collide. For each case, LiDARCrafter faithfully generates point cloud sequences that precisely match the specified layout and maintain strong temporal consistency, while preserving the geometric features of all foreground objects. These results demonstrate the model's powerful capability for creative, controllable corner case generation, which is highly valuable for evaluating and improving the safety performance of downstream perception and planning models.

\begin{figure*}[t]
    \centering
    \includegraphics[width=\textwidth]{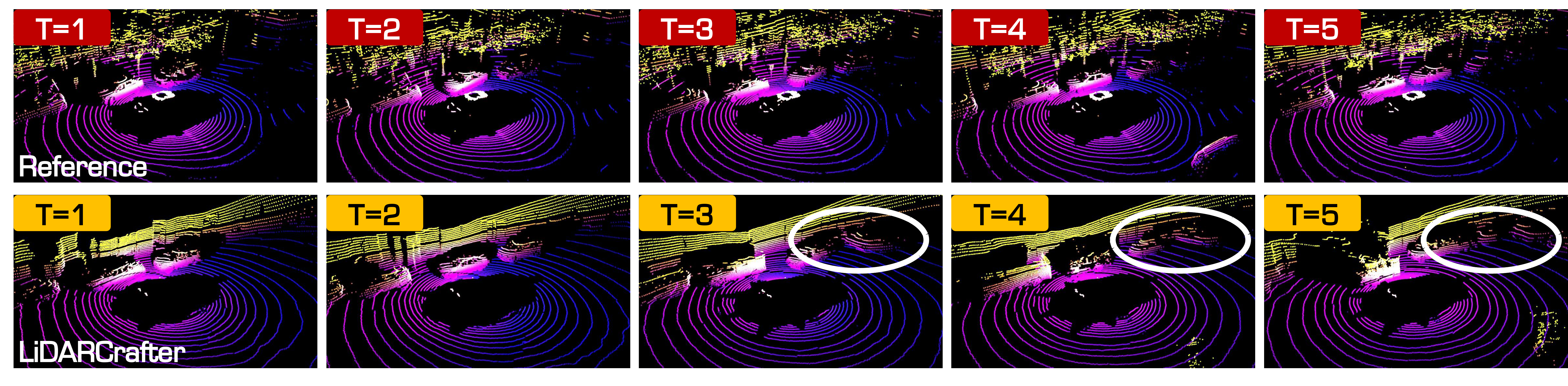}
    \vspace{-0.7cm}
    \caption{
        Example failure case: an initially sharp vehicle (left) becomes blurred in the third frame due to error accumulation in autoregressive generation, resulting in persistent geometric distortion in later frames. This highlights the need to explicitly incorporate historical geometric context (e.g., through inpainting) to enhance long-term consistency.
    }
    \label{supp_fig:failure_case}
\end{figure*}

\subsection{Failure Case Examples}
Despite the strong balance between temporal consistency and fidelity achieved by LiDARCrafter, our method still encounters several failure modes. As shown in Figure~\ref{supp_fig:failure_case}, the autoregressive, frame-by-frame generation paradigm introduces a dependence on the quality of previous frames, resulting in the potential for error accumulation over time. In the illustrated case, a vehicle becomes blurred in the third frame, leading to persistent geometric distortion and positional ambiguity in subsequent frames. 

We attribute such issues to the inherent stochasticity of generating each frame from pure noise, which may occasionally diverge from plausible scene evolution. This observation suggests a promising future direction: explicitly integrating geometric information from historical frames into the generation process. By adopting inpainting-like mechanisms, the model could better preserve accurate object geometry and mitigate error accumulation in temporally evolving scenes.

%% file: supp_sections/3_impact.tex
\section{Broader Impact}
In this section, we discuss the broader societal implications, potential limitations, and future research directions associated with our proposed LiDARCrafter framework for 4D LiDAR sequence generation.

\subsection{Potential Societal Impact}
The ability to controllably generate realistic and diverse 4D LiDAR data can substantially benefit autonomous driving research and deployment. Our framework enables scalable simulation of rare or hazardous driving scenarios, facilitating rigorous validation of perception and planning algorithms. This, in turn, may accelerate the safe deployment of intelligent vehicles, reduce reliance on costly data collection and manual annotation, and enhance the robustness of models in real world environments. However, care should be taken to prevent misuse, such as generating misleading or falsified data for safety critical applications.

\subsection{Potential Limitations}
Despite the advantages demonstrated in this work, several limitations remain. LiDARCrafter currently focuses on scenarios represented in the nuScenes dataset, and its generalizability to other sensor configurations or highly complex urban environments is yet to be explored. While our autoregressive design enables high-quality temporal synthesis, it may suffer from error accumulation and mode collapse in long sequences, as discussed in our failure case analysis. Moreover, the conditioning mechanisms while expressive may not cover all possible user intents or semantic constraints, potentially limiting fine-grained controllability in highly heterogeneous scenes.

\subsection{Future Directions}
Our experiments reveal a fundamental trade-off between LiDAR point cloud representations. Voxel-based approaches exhibit a natural advantage for maintaining temporal consistency, as their 3D discretization avoids the scale inconsistency present in range-view representations and is amenable to a variety of conditioning signals. However, voxelization tends to disrupt the raw LiDAR scanning pattern, diminishing point cloud fidelity. In contrast, range-based representations faithfully preserve LiDAR patterns and scene geometry, but embedding complex conditioning information remains challenging.

In future work, we aim to bridge the gap between these two paradigms by developing unified frameworks that align range and voxel representations. Such approaches could leverage the fidelity of range images while exploiting the temporal and conditional flexibility of voxel-based methods. By combining the strengths of both representations, we hope to further enhance the controllability and realism of generated 4D LiDAR data, expanding their utility for downstream autonomous driving applications and simulation.

%% file: supp_sections/4_ack.tex
\section{Public Resources Used}
\label{sec:public-resources-used}

In this section, we acknowledge the use of the following public resources, during the course of this work.

\subsection{Public Codebase Used}
We acknowledge the use of the following public codebase, during the course of this work:
\subsection{Public Codebase Used}
We acknowledge the use of the following public codebase, during the course of this work:
\begin{itemize}
    \item MMEngine\footnote{\url{https://github.com/open-mmlab/mmengine}.} \dotfill Apache License 2.0
    
    \item MMCV\footnote{\url{https://github.com/open-mmlab/mmcv}.} \dotfill Apache License 2.0
    
    \item MMDetection\footnote{\url{https://github.com/open-mmlab/mmdetection}.} \dotfill Apache License 2.0
    
    \item MMDetection3D\footnote{\url{https://github.com/open-mmlab/mmdetection3d}.} \dotfill Apache License 2.0
    
    \item OpenPCSeg\footnote{\url{https://github.com/PJLab-ADG/OpenPCSeg}.} \dotfill Apache License 2.0
    \item OpenPCDet\footnote{\url{https://github.com/open-mmlab/OpenPCDet}.} \dotfill Apache License 2.0
\end{itemize}

\subsection{Public Datasets Used}
We acknowledge the use of the following public datasets, during the course of this work:
\begin{itemize}
    \item nuScenes\footnote{\url{https://www.nuscenes.org/nuscenes}.} \dotfill CC BY-NC-SA 4.0
\end{itemize}

\subsection{Public Implementations Used}
\begin{itemize}
    \item nuscenes-devkit\footnote{\url{https://github.com/nutonomy/nuscenes-devkit}.} \dotfill Apache License 2.0

    \item Open3D\footnote{\url{http://www.open3d.org}.} \dotfill MIT License  
    
    \item PyTorch\footnote{\url{https://pytorch.org}.} \dotfill BSD License  
    
    \item TorchSparse\footnote{\url{https://github.com/mit-han-lab/torchsparse}.} \dotfill MIT License
\end{itemize}